\documentclass{article}
\pdfoutput=1

\usepackage{arxiv}

\usepackage[utf8]{inputenc} % allow utf-8 input
\usepackage[T1]{fontenc}    % use 8-bit T1 fonts
\usepackage{hyperref}       % hyperlinks
\usepackage{url}            % simple URL typesetting
\usepackage{booktabs}       % professional-quality tables
\usepackage{amsfonts}       % blackboard math symbols
\usepackage{nicefrac}       % compact symbols for 1/2, etc.
\usepackage{microtype}      % microtypography

\usepackage{times}
\usepackage{epsfig}
\usepackage{graphicx}
\usepackage{amsmath}
\usepackage{amssymb}
\usepackage[titletoc]{appendix}

\usepackage{caption} 
\usepackage{color} 
\graphicspath{ {./images/} }
\usepackage{subcaption}
\usepackage{amsmath} 
\usepackage{multirow}
\usepackage{xcolor}
\usepackage{lipsum}

\title{Edge-Guided Occlusion Fading Reduction for a Light-Weighted Self-Supervised Monocular Depth Estimation}

\author{Kuo-Shiuan Peng, Gregory Ditzler and Jerzy Rozenblit\\
  University of Arizona\\
  Tucson, AZ 85721\\
  {\tt\small \{kspeng, ditzler, jerzyr\}@email.arizona.edu}
}
\begin{document}
\maketitle

\begin{abstract}
Self-supervised monocular depth estimation methods generally suffer the occlusion fading issue due to the lack of supervision by the per pixel ground truth. Although a post-processing method was proposed by Godard et. al. \cite{Godard2017CVPR} to reduce the occlusion fading, the compensated results have a severe halo effect. In this paper, we propose a novel Edge-Guided post-processing to reduce the occlusion fading issue for self-supervised monocular depth estimation. We further introduce Atrous Spatial Pyramid Pooling (ASPP) into the network to reduce the computational costs and improve the inference performance. The proposed ASPP-based network is lighter, faster, and better than current commonly used depth estimation networks. This light-weight network only needs 8.1 million parameters and can achieve up to 40 frames per second for $256\times512$ input in the inference stage using a single nVIDIA GTX1080 GPU. The proposed network also outperforms the current state-of-the-art on the KITTI benchmarks. The ASPP-based network and Edge-Guided post-processing produce better results either quantitatively and qualitatively than the competitors.

\end{abstract}

% keywords can be removed

\section{Introduction}

Depth estimation is one of the fundamental problems with a long history in computer vision. It also serves as the cornerstone for many machine perception applications, such as 3D reconstruction, auto-driving system, industrial machine vision, robotics interaction, etc. However, most research is performed based on the availability of multiple observations in target scenes. The constraint of the multiple observations can be overcome by the supervised method because of the emerging deep learning technology \cite{Ladicky2014CVPR, Eigen2014NIPS, Liu2016PAMI}. These methods aim to directly predict the pixel depth from a single image by learning the given ground truth depth data of a large amount of dataset. Despite the promising results of the monocular depth prediction, these methods suffer from the limitation of the quality and availability of the collected ground truth pixel depth. Hence, the self-supervised approaches learning the depth information from a single image has received increasing attention in recent research.

\begin{figure}[ht]
\begin{center}
	\includegraphics[width=0.450\textwidth]{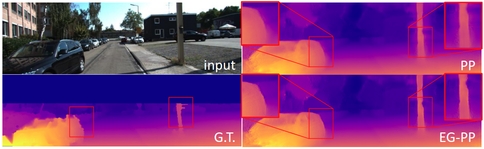}
\end{center}	
\caption{Comparison between the conventional post-processing (PP) \cite{Godard2017CVPR} and the proposed Edge-Guided post-processing (EG-PP) on KITTI dataset. Our method can reserve the sharp edge of the detected object depth and avoid the halo effect. }
\label{fig:lw-eg-mde-demo}
\end{figure}

In the task of monocular depth estimation, the input source is a monocular image (e.g., a left image) and the stereo pairs are available during training. Then the corresponding another view (e.g., the right image), can be reconstructed by the estimated right depth and the input left image (left) using a warping function \cite{Max2015NIPS}. Hence, the reconstructed right view is supervised by an actual right image. The  estimated depth can also be calibrated in the regression for the reconstructed right view. One of the major challenges of the self-supervision method is to reduce false detection using a compact network. It's been shown in \cite{Godard2017CVPR} that the deeper network (e.g. Resnet50) can yield better estimated depth compared to a compact network (e.g. VGG19). However, very deep networks are inefficient for real-time usage. A high performance light-weight network design for a depth estimation network is needed. Another challenge is the stereo dis-occlusion effect. Disparity ramps happen in the stereo dis-occlusion area of the estimated disparity and largely downgrade the estimation quality quantitatively and qualitatively. Although a post-processing method is proposed to compensate the occlusion shading using a flip prediction alignment method \cite{Godard2017CVPR}, the compensated output suffers severe halo effect as shown in Fig. \ref{fig:lw-eg-mde-demo}(PP). A solution of solving this issue requires high quality estimation results. 
%In this paper, we propose a light-weight depth estimation network to improve either the estimation performance or computation costs. Furthermore, an Edge-guided post-processing is presented to resolve the occlusion shading issue with minimal halo effects as shown in Fig. \ref{fig:lw-eg-mde-demo}. 

%%%% seems redundant with what is below 
%In our experiments, we show that our method has superior design in network structure than the recent works and the proposed Edge-Guided post-processing can boost the estimation performance. The computational costs, including parameters and computation time of the proposed method, are much lower than the current existing method. Per prediction only needs around 25ms (40FPS). We have much lower false detection of the objects and can generate more clear object shapes.   

The main contributions of this paper are as follows: (1) We propose a Light-weight DispNet that is smaller, faster, and better than the conventional DispNet. We also point out that the last two dense feature layers of the encoder in DispNet are less efficient in extracting long-range features. 
(2) A novel Edge-Guided post-processing method is presented in this paper. The occlusion fading is largely reduced with a minimized halo effect after applying our method. This Edge-Guided post-processing is universal and can be applied to any other self-supervised method.
(3) We evaluate our approach compared to the state-of-the-art on the challenging KITTI dataset. We test both the intrinsic network and integrated performance separately to demonstrate that our method has fundamentally improved the network performance. In the integration test, we showcase that our results can compete with the method that uses the semantic consistency supervision. 

%\begin{enumerate}
%\item We propose a Light-weight DispNet that is smaller, faster, and better than the conventional DispNet. We also point out that the last two dense feature layers of the encoder in DispNet are less efficient in extracting long-range features. 
%\item A novel Edge-Guided post-processing method is presented in this paper. The occlusion fading is largely reduced with a minimized halo effect after applying our method. This Edge-Guided post-processing is universal and can be applied to any other self-supervised method.
%\item We evaluate our approach compared to the state-of-the-art on the challenging KITTI dataset. We test both the intrinsic network and integrated performance separately to demonstrate that our method has fundamentally improved the network performance. In the integration test, we showcase that our results can compete with the method that uses the semantic consistency supervision.   
%\end{enumerate}

\section{Related Works}

A depth map is a form of an absolute depth or disparity value. The depth is inversely proportional to the disparity. The former and latter values can be converted to each other based on the parameters of the rectified multi-cameras. We use monocular depth estimation because they use a single monocular image as the input to the system rather than using multi-view images to calculate the disparity of the scene \cite{Ting2016cvpr}. In \cite{Liu2016PAMI},  CNNs and continuous CRFs are used as a patch-wise depth predictor to estimate the depth information in a supervised manner, which required a large amount of high-quality ground truth images. A combination of coarse and fine cues was proposed to improve the performance of the depth estimation but still was supervised \cite{David2014nips}. On the other hand, self-supervised approaches only need supervision from either stereo image pairs \cite{Garg2016ECCV, Godard2017CVPR, Alex2019cvpr} or monocular video frames \cite{Hyowon2016cvpr, Tinghui2017cvpr, Sudheendra2017arXiv, Reza2018cvpr, Zhichao2018cvpr, Huangying2018cvpr}. In this approach, the disparity is used as an intermediate product which can be converted to reconstruct the images with the inverse warping transform \cite{Max2015NIPS}. To improve the performance of the self-supervised algorithm, various objective functions are proposed, such as left-right consistency \cite{Godard2017CVPR}, correlational consistency \cite{Kuo2019ICISIP}, and adaptive global and local error \cite{Alex2019cvpr}.

Self-supervised approaches have limited predictive abilities on finer details in an image compared to their supervised counterparts. There were different methods devised to mitigate this limitation. A semi-supervised approach is a combination of self-supervised and supervised methods \cite{Yevhen2017cvpr, Ali2019arXiv} . These approaches predict the dense pixel depth on pixels with ground truth from the supervised information and the pixels without ground truth in the self-supervised approach. These methods were particularly useful when the ground truth was sparse, and they yield promising results. Rather than using ground truth information, a joint prediction approach can use the semantic segmentation to assist in depth estimation \cite{po2019cvpr}. In semi-supervised approaches, the implicit methods were used to predict the depth and semantic segments jointly to leverage the common feature representations between two tasks; however, each task used independent objective functions \cite{Lubor2014cvpr, David2015iccv, Peng2015cvpr, Arsalan20163dv}. More recent work introduced the depth-semantic gradient consistency explicitly to refine the fine details of the estimated depth map \cite{Jianbo2018eccv, Pierluigi2018accv, Zhenyu2018eccv, po2019cvpr}. 

One limitation of self-supervision method was the stereo dis-occlusion effect. The self-supervision method relied on stereo image pairs to calibrate the estimation without the ground truth. This method inherits the stereo dis-occlusion effect from the objective function that uses stereo image pairs. The early researches in the occlusion detection used Bayesian formulation \cite{Eric2003eai}, AdaBoost \cite{Ahmad2005tog}, and Random Forest based framework \cite{Ahmad2011cvpr} to detect the motion occlusion regions of motion images. These priors handcrafted the features to proceed the machine learning algorithms. Recently, a learning-based method has considered structural left-right symmetry \cite{Ang2018arXiv} to detect occlusions, while a\cite{Eddy2018eccv} proposed to estimate occlusions, motion, and depth boundary using a single network. These prior had shown plausible results but need a complicated model to only detect the occlusion. Recently, \cite{Godard2017CVPR} proposed to use a simple post processing module of flip prediction alignment method to recover the information of the occlusion area for the self-supervision approaches. Unfortunately, halo effects occur after applying this post processing module to the estimation.  None of the current methods can fit the needs of reducing the occlusion fading for a self-supervised depth estimation task.

In this paper, we  improve the depth network efficiency and performance quantitatively and qualitatively. However, there are only a few works that focus on optimizing the network structure for self-supervision in real-time. A Light-Weight RefineNet was proposed for joint semantic segmentation and depth estimation task \cite{Vladimir2018arxiv}. This methods is designed for the supervision method. We have tested it and found out that its performance is limited when applying to the self-supervision method. When we studied the multi-task network, we realized that the depth estimation and semantic segmentation can share the same feature representation in the network. Based on this finding, we argue that the semantic segmentation network structure can be used in the depth estimation network. In this paper, we introduce Atrous Spatial Pyramid Pooling (ASPP) module into our depth network from Deeplab semantic segmentation network \cite{Liang2017arxiv}. We successfully designed a Light-weight DispNet that has only $20\%$ size of the conventional depth network but is about $35\%$ faster in prediction. To address the occluding fading issue, we proposed an Edge-Guided post-processing method using Tensorflow right after the depth network to eliminate the occluding fading and halo effects in inference stage. The proposed method effectively improve both the quantitative and qualitative results and is suitable for all the self-supervision-based methods. 

\section{Methodology}

Our model is inspired by the works of \cite{Liang2017arxiv} and \cite{Godard2017CVPR}. We first introduce the ASPP module \cite{Liang2017arxiv} into our network design, and then the objective function is directly adopted from \cite{Godard2017CVPR}. The proposed Edge-Guided post-processing is explained in the last section.

\subsection{Light-Weight Disparity Network}

Many recent works designed their network by referring to DispNet \cite{Mayer2016cvpr}, which is an autoencoder-based architecture. The multi-scale features are exploited from the encoder, and the spatial resolution is recovered from the decoder. The recovered multi-scale spatial resolutions are the estimated disparities. 

Since it has been shown that the depth estimation and semantic segmentation prediction have common feature representations and they can share the base-network to perform multi-task prediction \cite{Vladimir2019ICRA}, we use the network design concept of the semantic segmentation task. It indicates that an extra module is commonly cascaded on top of the original network for detecting long-range information \cite{Liang2017arxiv, Liang2018ECCV}. We follow this design rule to modify the DispNet to do the depth estimation. We choose the ASPP module \cite{Liang2017arxiv} as our long-range detection module and insert it between the encoder and decoder. 

\begin{figure}[t]
\begin{center}
	\includegraphics[width=0.4\textwidth]{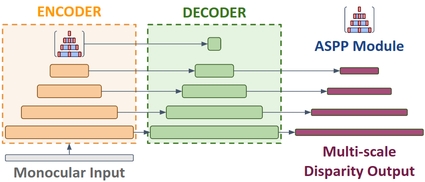}
\end{center}	
\caption{Light-Weight DispNet Structure.}
\label{fig:lw-eg-network}
\end{figure}

To further optimize our network, we analyzed the feature layers of the encoder, and we found that the last few dense feature layers have a minor contribution to the estimation, especially after introducing the ASPP module. Based on this observation, we simplify the DispNet by using ASPP module to replace the last two dense layers of the encoder. This design successfully reduces the network size of the network and produce a better performance. We name this structure a Light-Weight DispNet. The proposed network structure is shown in Fig. \ref{fig:lw-eg-network}. We here use \cite{Godard2017CVPR} as a baseline example. If DispNet uses VGG19 as the backbone, the network parameters are about 31.6 million, and the inference time is about 32 msec. The corresponding Light-Weight DispNet only need 8.1 million ($74\%$ less) with inference time 25.2 msec ($22\%$ less). Nevertheless, the proposed Light-Weight DispNet inherits the trait of DispNet that the backbone of the encoder can be modified. In this paper, we choose VGG8 and Resnet24 as our backbones. We demonstrate that the ASPP can effectively improve the estimation performance in our evaluation. 
Please refer to Appendix for the detailed network design.
%We present the brief network structures of proposed VGGASPP and ResASPP in Table \ref{tbl:lw-eg-vggaspp} and \ref{tbl:lw-eg-resaspp}. $^\star$Please refer to the supplemental material for the detailed network design.

\subsection{Objective Function}

We decide to adopt the objective function from \cite{Godard2017CVPR} directly with several reasons. The most important consideration is that the aim of the left-right consistency function from \cite{Godard2017CVPR} has demonstrated promising results among the recent works. The successors only have minor modifications. Besides, we would like to showcase that the proposed Light-Weight DispNet is substantially better than the conventional DispNet using the same objective function. 

The objective function is a weighted sum of three terms: appearance $(C_{ap})$, disparity smoothness $(C_{ds})$, and left-right consistency $(C_{cor})$. The self-supervise total loss is defined as following:
\begin{equation}
C_s = \alpha_{ap} \times C_{ap} + \alpha_{ds} \times C_{ds} +  \alpha_{lr}\times C_{cor}
\label{eq:ss-ss-total-loss}
\end{equation}
The weights $(\alpha_{ap}, \alpha_{ds}, \alpha_{lr})$ are determined before optimization and set as $(1.0, 0.5, 1.0)$. The definition of each term can be found in \cite{Godard2017CVPR}.
\begin{figure*}
\begin{center}
	\includegraphics[width=0.90\textwidth]{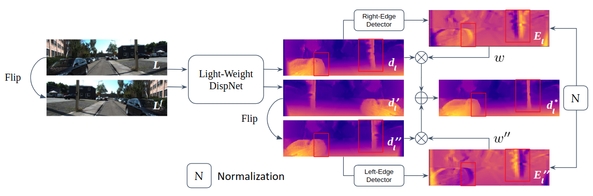}
\end{center}	
\caption{Edge-Guided Post Processing}
\label{fig:lw-eg-post-proc}
\end{figure*}

\subsection{Edge-Guided Post-Processing}

The stereo dis-occlusion effect is one of the limitations in the self-supervision method of monocular depth estimation. Stereo dis-occlusion creates disparity ramps (occlusion fading) on both the left side of the image and the occluders. \cite{Godard2017CVPR} proposed a post-processing method to reduce this effect. This post-processing estimates the disparity map $d_l$ and the flipped disparity map  $d'_l$, which are from input image $I$ and its horizontally flipped image $I'$. Then the flipped disparity map $d'_l$ is flipped back as a $d''_l$ that aligns with  $d_l$ but where the occlusion fading is on the right of occluders as well as on the right side of the image. The final result is an average of $d_l$ and $d''_l$ but assigning the first $5\%$ on the left of the image using $d_l$ and the last $5\%$ on the right to the disparities from $d_l$. The $5\%$ of left and right original disparities is the reserved boundary range to avoid the boundary fading during the disparity synthesis in the post processing.

This post-processing uses a mirror trick to generate a well-aligned projected disparity $d''_l$ that has right-side occlusion fading. The average of $d_l$ and $d''_l$ can reduce the left-side occlusion fading because $d''_l$ has correct left-side estimation results. However, the right-side occlusion fading is also being involved. This average process in the post-processing causes the halo effect in the final results, as shown in PP of Fig \ref{fig:lw-eg-mde-demo}. Instead of average, we propose an Edge-Guided weighted sum to suppress the occlusion fading of both $d_l$ and $d''_l$ in the combination to reduce the halo effect, as shown in EG-PP of Fig \ref{fig:lw-eg-mde-demo}. 

The proposed Edge-Guided post-processing is depicted in Fig. \ref{fig:lw-eg-post-proc}. We follow the design concept of \cite{Godard2017CVPR} to compute $d_l$ and $d''_l$, but we add edge-aware weights ($w$, $w''$) in the final combination. Here we take $w$ as an example to illustrate the algorithm. A right-edge detector is designed to extract the regional-edge confidence $E$. Instead of using Sobel detector, a wide-range horizontal gradient filter ($f_{gx}$) is used:

\begin{equation}
f_{gx}=\begin{bmatrix}
1 & ... & 0 & -1 & ...\\ 
1 & ... & 0 & -1 & ...\\ 
1 & ... & 0 & -1 & ... 
\end{bmatrix}_{3\times(2N)}/(3\times(2N))
\label{eq:lw-eg-fgx}
\end{equation}
where $N$ is the detection radius, whose default value is set to 10. After the convolution process $(\otimes)$ of $d_l$ and $f_{gx}$, we add an offset ($b$) and a gain ($a$) on the convolution result. Then a sigmoid function is applied:

\begin{equation}
E_l = \text{sigmoid}((d_{l}\otimes{f_{gx}}-b)*a)
\label{eq:lw-eg-el}
\end{equation}
where $E_l$ is the right regional-edge confidence. The offset $b$ and gain $a$ are set as 0.5 and 32 to maximize the $E_l$ in the range [0, 1]. In this equation, the right edge region has the confidence close to 1, while the left occlusion fading area has the confidence close to 0. The confidence of the flat area keeps around 0.5. The $E''_l$ is obtained in the same way but using the horizontal flipped $f_{gx}$ as the left-edge detector. The last step is to normalize $E_l$ and $E''_l$ to obtain $w$ and $w''$. Then the final output $d^{\star}_l$ is a weighted sum of $d_l$ and $d''_l$:

\begin{equation}
\begin{aligned}
 &w = E_l / (E_l+E''_l) \\
 &w'' = E''_l / (E_l+E''_l)
\label{eq:lw-eg-w}
\end{aligned}
\end{equation}

\begin{equation}
d^{\star}_l = wd_l + w''d''_l
\label{eq:lw-eg-dout}
\end{equation}
The normalization process is needed to prevent overlap detection between $E_l$ and $E''_l$. It ensures that the sum of $w$ and $w''$ is 1 for each pixel and the final output $d^{\star}_l$ has no overhead compared to $d_l$ and $d''_l$. 
%We implement the proposed Edge-Guided post-processing using Tensorflow \cite{tensorflow2015-whitepaper}. 
There are no learning parameters and the computation cost is very low. % and less than $4\%$ computation time is needed in the inference stage. 
We also follow the boundary reserving design here with the $2\%$ reserved boundary range. The details are described in Appendix.

\section{Experiments}

We compare the performance of our approach to recent self-supervised monocular depth estimation methods. We selected Godard et al. \cite{Godard2017CVPR}'s work as our baseline and used the same benchmark configurations \cite{Godard2017CVPR}. We evaluate our approach in multiple aspects of KITTI \cite{Geiger2012cvpr} dataset quantitatively and qualitatively. The ablation study is first conducted to prove the effectiveness of our approach using KITTI split. We then have a benchmark with the current start-of-the-art on Eigen split. We split the benchmark into two parts: network and integration test. In the network test, we showcase that the proposed network structure is better than the competitors without any post-processing. The integration test is to demonstrate the power of the proposed Edge-Guided post-processing method. We involve the semantic consistency of semi-supervision learning \cite{po2019cvpr}. %We show that our result can compete with priors that even include the semantic consistency factor. \
Please refer to our code \footnote{Available at \href {https://github.com/kspeng/lw-eg-monodepth}{github.com/kspeng/lw-eg-monodepth}} to see the detailed implementation.

\subsection{Datasets}

We evaluate the performance of the proposed method on the KITTI benchmark \cite{Geiger2012cvpr}. We use two different test splits, KITTI and Eigen Split \cite{Eigen2014NIPS}, of KITTI dataset to do the ablation analysis of the variants of our method and the benchmark compared with the existing works. We follow the approach of  \cite{Godard2017CVPR} to use 29,000 image pairs as the training set.
%The KITTI dataset contains 42,382 rectified stereo pairs in raw form from 61 scenes. 
The typical image size of the KITTI dataset is $375\times1242$ pixels, and the stereo image pairs are well-calibrated in the calibrated camera configuration. 
The depth labels were collected from a Velodyne laser sensor. 
%and in the form of a sparse 3D laser measurement. 
%The parameter of the stereo setup is clearly defined in the dataset as well, so the predicted disparities can be converted back to the corresponding depth. 

Furthermore, it has been shown by \cite{Godard2017CVPR} that pre-training with Cityscapes dataset \cite{Marius2016cvpr} can improve the performance on KITTI benchmark. We also include this strategy in the benchmark. Cityscapes dataset contains 22,973 higher resolution and image quality training stereo pairs. We use the same crop as \cite{Godard2017CVPR} to avoid the reflective car hoods from the input. In the combinational training on Cityscapes and KITTI dataset (C+K), we pre-train our models with an eight batches and 50 epochs initial training on Cityscapes dataset, and then another eight batches and 100 epochs training on KITTI dataset. 
\subsection{Metrics}

We use the evaluation metrics from \cite{Geiger2012cvpr} for depth estimation, which measure the error in meters from the ground truth and the percentage of depth that is within a threshold from the correct value. The error measurements represent the average error. 
%Hence, the lower value is better. The percentage of the correct value is an alias of accuracy, and higher valuer is better.  

\subsection{Implementations}

\paragraph{Configuration} Our methods were implemented in Tensorflow 1.15 \cite{tensorflow2015-whitepaper} using Python 3.7 under the Ubuntu environment with a single NVIDIA GTX 1080 GPU. All input images are resized to $256\times512$ from the original size of the training image. All our models are trained by eight batches and 100 epochs on the KITTI training dataset. The predictions happen in around 25ms (around 40 frames per second (FPS)) of the proposed  VGGASPP model and 31ms (around 33 FPS) of the proposed ResASPP model.

\paragraph{Parameter Settings} We follow the prior work  \cite{Godard2017CVPR} to set up the training optimizer and parameters. We train our models from scratch and use the Adam optimizer. The  training flow involves a batch size 8 and 100 epochs. The model converges after 80 epochs, and the improvement after 100 epochs was minor. 
%The training time for VGGASPP and ResASPP is about 24 and 33 hours. 
%%%% mention this in the appendix 
%We start from the initial learning rate of $\lambda = 10^{-4} $ and decrease $\lambda$ by half every ten epochs. During the training, the dataset is randomly augmented, including horizontal flipping, color/gamma/brightness adjustment by 50\% chance. 

\paragraph{Network Backbone} In the evaluation, we show the experimental results in two models, VGGASPP (VGG8 Backbone) and ResASPP (Resnet24 Backbone), of our Light-weight DispNet. The computation costs of each backbone are also summarized in the Ablation section. We show that both our models have better performance than competitors in the benchmark studies.

\subsection{Results}

\subsubsection{Ablation Study}

In the ablation study, we analyze the quantitative performance improvement and the computational costs of our various designs using KITTI split on the KITTI dataset. For the quantitative performance improvement, we use two models, VGG and Resnet50, of our prior work \cite{Godard2017CVPR} without post-processing as the baseline. The results are shown in Table \ref{tbl:ablation-nums}. We can see that our VGGASPP without conventional post-processing outperforms the VGG model of the baseline and is very close to the Resnet50 one. On the other hand, our ResASPP without conventional post-processing has better performance than both VGG and Resnet50 models on the baseline. Furthermore, compared to the conventional post-processing, the proposed Edge-Guided post-processing has a significant performance boost on both our two models, especially on terms of RMSE(log) and $\delta<1.25$, which are the most challenging parts. We also add the C+K training set up in the last section. We found that our VGGASPP can even outperform ResASPP using the Edge-Guided post-processing.

\begin{table*}[t]
\centering
\resizebox{\textwidth}{!}{\begin{tabular}{lccccccccccc}
    \hline
\multicolumn{1}{c}{\multirow{2}{*}{Approach}} & 
\multicolumn{1}{c}{\multirow{2}{*}{Train}} & 
\multicolumn{1}{c}{\multirow{2}{*}{PP}} & 
\multicolumn{1}{l}{ARD} & \multicolumn{1}{l}{SRD} & \multicolumn{1}{l}{RMSE} & \multicolumn{1}{l}{RMSE(log)}  & \multicolumn{1}{l}{D1-all}  &  & \multicolumn{1}{l}{$\delta<1.25$}  & \multicolumn{1}{l}{$\delta<1.25^2$}  & \multicolumn{1}{l}{$\delta<1.25^3$} \\ 
\cline{4-8} \cline{10-12} \multicolumn{3}{c}{} & \multicolumn{5}{c}{Lower is better.} &  & \multicolumn{3}{c}{Higher is better.} \\  
    \hline
Godard et al. \cite{Godard2017CVPR} VGG & K & N & 0.1240 & 1.3880 & 6.125 & 0.217 & 30.272 & & 0.841 & 0.936 & 0.975 \\
Godard et al. \cite{Godard2017CVPR} Resnet50 & K & N & 0.1127 & 1.1331 & 5.749 & 0.203 & 29.553 & & 0.851 & 0.947 & 0.980 \\
    \hline
Our VGGASPP & K & N & 0.1134 & 1.1636 & 5.734 & 0.201 & 27.379 & & 0.853 & 0.945 & 0.979 \\
Our VGGASPP & K & Y & 0.1079 & 1.0259 & 5.464 & 0.192 & 26.395 & & 0.857 & 0.949 & 0.982 \\
Our VGGASPP & K & Y+ & 0.1077 & 1.0238 & 5.387 & 0.189 & 26.152 & & 0.860 & 0.951 & 0.983 \\
Our ResASPP & K & N & 0.1107 & 1.0633 & 5.612 & 0.199 & 27.531 & & 0.854 & 0.946 & 0.980 \\
Our ResASPP & K & Y & 0.1075 & 0.9878 & 5.474 & 0.193 & 27.050 & & 0.855 & 0.949 & 0.981 \\
Our ResASPP & K & Y+ & 0.1071 & 0.9936 & 5.394 & 0.189 & 26.673 & & 0.858 & 0.952 & 0.983 \\
    \hline
\iffalse
Godard et al. \cite{Godard2017CVPR} VGG & C+K & N 
 & 0.104 & 1.070 & 5.417 & 0.188 & 25.523 & & 0.875 & 0.956 & 0.983 \\
Godard et al. \cite{Godard2017CVPR} VGG & C+K & Y & 0.100 & 0.934 & 5.141 & 0.178 & 25.077 & & 0.878 & 0.961 & 0.986 \\
Godard et al. \cite{Godard2017CVPR} Resnet50 & C+K & Y & \textbf{0.097} & 
\textbf{0.896} & 5.093 & 0.176 & 23.811 & & 0.879 & \textbf{0.962} & 0.986 \\ 
Our VGGASPP & C+K & Y & 0.0987 & 0.9130 & 5.129 & 0.179 & 23.361 & & 0.879 & 0.958 & 0.985 Our ResASPP & C+K & Y & 0.1003 & 0.9671 & 5.160 & 0.179 & 23.633 & & 0.875 & 0.958 & 0.986 \\
\\
\fi
Our VGGASPP & C+K & Y+ & \textbf{0.0984} & \textbf{0.9196} & \textbf{5.035} & \textbf{0.175} & \textbf{22.942} & & \textbf{0.883} & \textbf{0.961} & 0.986 \\
Our ResASPP & C+K & Y+ & 0.1000 & 0.9697 & 5.070 & 0.175 & 23.231 & & 0.879 & \textbf{0.961} & \textbf{0.987} \\
    \hline
\end{tabular}}
\caption{Quantitative results for different variants of our approach on the KITTI Stereo 2015 test dataset. The PP means using post-processing. N means no PP, Y means conventional PP, and Y+ represents the proposed Edge-Guided PP. The best result in each subsection is shown in bold. The training scenario is based on the KITTI training set (K), while the last section shows the results which are pre-trained by Cityscapes training sets (C+K). We use our prior \cite{Godard2017CVPR} as our baseline is shown in the first section. 
} 
\label{tbl:ablation-nums}
\end{table*}

In Table \ref{tbl:ablation-costs}, we can see the computation costs of both our methods with the baseline. Our VGGASPP has only $25.6\%$ of parameters but is $23.2\%$ faster in the prediction compared to VGG model of the baseline. Our ResASPP has only $13.8\%$ of parameters but performs $48\%$ more quickly in the prediction compared to the Resnet50 model of the baseline. Only $3.8\%$ and $3\%$ loss in computation time occur in prediction on VGGASPP and ResASPP models when we apply the Edge-Guided post-processing to our methods. There is no need for extra parameters. 

\begin{table}[!t]
\centering
\resizebox{0.45\linewidth}{!}{\begin{tabular}{lccc}
\hline
\multicolumn{1}{c}{Approach} & Parameters & Predict(ms/FPS) \\ 
	\hline
Godard et al. \cite{Godard2017CVPR} VGG      & 31600072 & 31/32.2 \\
Godard et al. \cite{Godard2017CVPR} Resnet50 & 58452008 & 44/22.7 \\
	\hline
Our VGGASPP & \textbf{8134344} & \textbf{25.2/39.7} \\
Our ResASPP & 11825448 & 29.7/33.6 \\
Our VGGASPP+EGPP & \textbf{8134344}  & 26.2/38.2 \\
Our ResASPP+EGPP & 11825448 & 30.6/32.6 \\
    \hline
\end{tabular}}
\caption{Computational costs of different variants of our approach on the KITTI training dataset within 8 batches and 100 epochs. The units of training of prediction are msec(ms) and frame per second(FPS). Smaller parameters and ms values represent lower cost. %Best results shown in bold.
}
\label{tbl:ablation-costs}
\end{table}

\subsubsection{State-of-the-art comparison}

\begin{table*}\small
\centering
\resizebox{\textwidth}{!}{\begin{tabular}{lccccccccccc}
    \hline
\multicolumn{1}{c}{\multirow{2}{*}{Approach}} &
\multicolumn{1}{c}{\multirow{2}{*}{Train}} & 
\multicolumn{1}{c}{\multirow{2}{*}{Test}} & 
\multicolumn{1}{c}{\multirow{2}{*}{PP}} & 
\multicolumn{1}{l}{ARD} & \multicolumn{1}{l}{SRD} & \multicolumn{1}{l}{RMSE} & \multicolumn{1}{l}{RMSE(log)}  &  & \multicolumn{1}{l}{$\delta<1.25$}  & \multicolumn{1}{l}{$\delta<1.25^2$}  & \multicolumn{1}{l}{$\delta<1.25^3$}  \\ 
\cline{5-8} \cline{10-12} \multicolumn{4}{c}{} & \multicolumn{4}{c}{Lower is better.} &  & \multicolumn{3}{c}{Higher is better.} \\  
    \hline
Zhou et al. \cite{Tinghui2017cvpr} & K & E - 80m & N & 0.2080 & 1.7680 & 6.8560 & 0.283 & & 0.678 & 0.885 & 0.957 \\
Mahjourian et al. \cite{Reza2018cvpr} & K & E - 80m & N & 0.1630 & 1.2400 & 6.2200 & 0.25 & & 0.762 & 0.916 & 0.968 \\
Garg et al. \cite{Garg2016ECCV}	& K & E - 80m & N & 0.1520 & 1.2260 & 5.849 & 0.246 & & 0.784 & 0.921 & 0.967 \\
Godard et al. \cite{Godard2017CVPR} & K & E - 80m & N & 0.1480 & 1.3440 & 5.927 & 0.247 & & 0.803 & 0.922 & 0.964 \\
Alex and Stefano \cite{Alex2019cvpr} & K & E - 80m & N & 0.1330 & 1.1260 & 5.515 & 0.231 & & 0.826 & 0.934 & 0.969 \\
Ours VGGASPP & K & E - 80m & N & 0.1146 & 1.0248 & 5.198 & 0.217 & & 0.851 & 0.936 & 0.970 \\
Ours ResASPP & K & E - 80m & N & \textbf{0.1124} & \textbf{1.0002} & \textbf{5.129} & \textbf{0.215} & & \textbf{0.855} & \textbf{0.938} & \textbf{0.971} \\ 
	\hline
Zhou et al. \cite{Tinghui2017cvpr}      & C+K & E - 80m & N & 0.198 & 1.836 & 6.565 & 0.275 & & 0.718 & 0.901 & 0.96 \\
Mahjourian et al. \cite{Reza2018cvpr}   & C+K & E - 80m & N & 0.159 & 1.231 & 5.912 & 0.243 & & 0.784 & 0.923 & 0.97 \\
Godard et al. \cite{Godard2017CVPR}  	& C+K & E - 80m & N & 0.1240 & 1.0760 & 5.311 & 0.219 & & 0.847 & 0.942 & 0.973 \\
Alex and Stefano \cite{Alex2019cvpr}	& C+K & E - 80m & N & 0.1180 & 0.9960 & 5.134 & 0.215 & & 0.849 & \textbf{0.945} & \textbf{0.975} \\
Ours VGGASPP & C+K & E - 50m & N & 0.1095 & 0.9405 & 5.001 & 0.208 & & 0.864 & 0.944 & 0.974 \\
Ours ResASPP & C+K & E - 50m & N & \textbf{0.1077} & \textbf{0.9303} & \textbf{4.954} & \textbf{0.208} & & \textbf{0.865} & 0.944 & 0.974 \\
	\hline
Garg et al. \cite{Godard2017CVPR}   & K & E - 50m & N & 0.1690 & 1.0800 & 5.1040 & 0.273 & & 0.740 & 0.904 & 0.962 \\
Zhou et al. \cite{Tinghui2017cvpr}  & K & E - 50m & N & 0.2010 & 1.3910 & 5.1810 & 0.264 & & 0.696 & 0.900 & 0.966 \\
Godard et al. \cite{Godard2017CVPR} & K & E - 50m & N & 0.1400 & 0.976 & 4.471 & 0.232 & & 0.818 & 0.931 & 0.969 \\
Alex and Stefano \cite{Alex2019cvpr} & K & E - 50m & N & 0.1260 & 0.832 & 4.172 & 0.217 & & 0.840 & 0.941 & 0.973 \\
Ours VGGASPP & K & E - 80m & N & 0.1077 & 0.7273 & 3.885 & 0.204 & & 0.865 & 0.944 & \textbf{0.974} \\
Ours ResASPP & K & E - 80m & N & 0.\textbf{1058} & \textbf{0.7211} & \textbf{3.846} & \textbf{0.202} & & \textbf{0.868} & \textbf{0.945} & \textbf{0.974} \\
	\hline
\end{tabular}}
\caption{Quantitative results of network test compared with the state-of-the-art on KITTI Eigen (E) split with KITTI (K) training set. For a fair comparison, we compare the reference methods without the post-processing (PP), and all the results use the crop defined by \cite{Garg2016ECCV}. We also examine the effects of the full distance (0-80m) and the near distance (1-50m). The results which are pre-trained with Cityscapes (C) are evaluated as well. %The best results are marked in bold. Our methods are superior to the reference methods in the benchmarks of different kinds of settings. 
}
\label{tbl:benchmark-network}
\end{table*}

\begin{figure*}[t]
\centering
	\begin{subfigure}[t]{.18\textwidth}
		\centering
		\includegraphics[width=.55\textwidth]{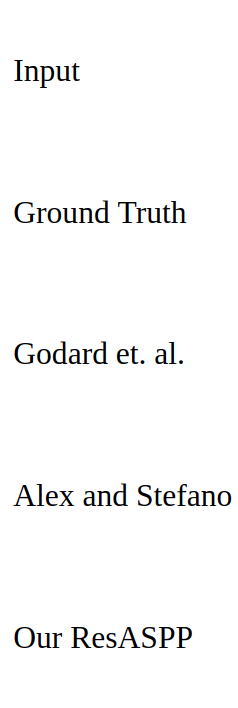}
	\end{subfigure}	
	\begin{subfigure}[t]{.2\textwidth}
		\centering
		\includegraphics[width=1.\textwidth]{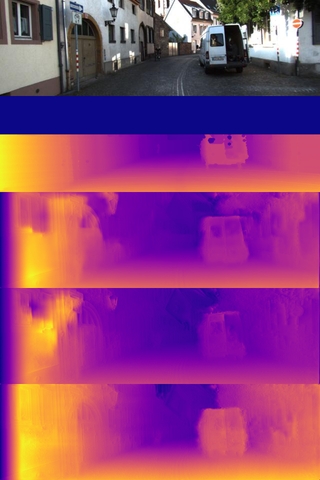}		
	\end{subfigure}
	\begin{subfigure}[t]{.2\textwidth}
		\centering
		\includegraphics[width=1.\textwidth]{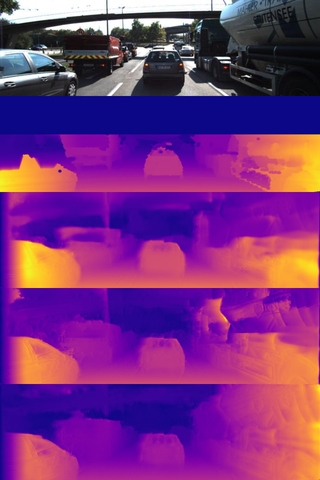} 
	\end{subfigure}
	\begin{subfigure}[t]{.2\textwidth}
		\centering
		\includegraphics[width=1.\textwidth]{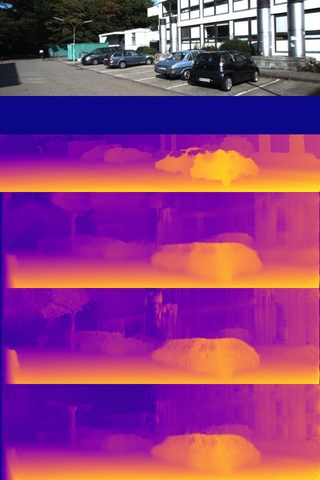}
	\end{subfigure}
	\begin{subfigure}[t]{.2\textwidth}
		\centering
		\includegraphics[width=1.\textwidth]{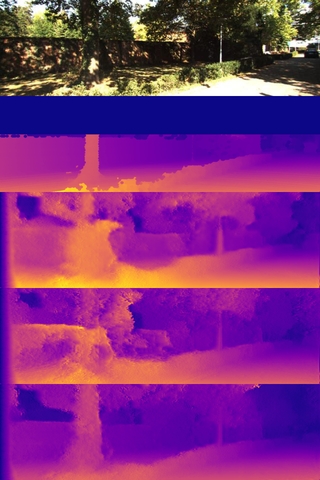}
	\end{subfigure}
\caption{Benchmark of qualitative results on KITTI dataset Eigen Split. 
%We interpolate the sparse ground truth velodyne depth for visualization. There are four sets in two parts listed in the Figure. 
%The images from top to down in each set are - source image, ground truth, Godard et al. \cite{Godard2018arxiv}, Alex and Stefano \cite{Alex2019cvpr}, and our ResASPP. 
All the results have no post-processing for a fair comparison. There are are many missing and false detection of trunks, poles, trucks, or cars in the reference methods. By contrast, our methods can accurately capture more fine objects and maintain their shapes better. }
\label{fig:demo-benchmark-network}
\end{figure*}

% In the benchmark, we have the network and integration test in separate tables and figures. 

\paragraph{Network Test} The network test considers the performance of the network without help from any post-processing methods in Table \ref{tbl:benchmark-network}. We involve several self-supervised priors, and the test is based on Eigen-slipt of KITTI dataset. The test is essentially in two types of examination, namely the results of the full distance (0-80m) and near distance (1-50m). We use the same crop defined by \cite{Garg2016ECCV} for a fair comparison. The training set is the KITTI dataset only, and we also add other C+K full distance results in the second section of the table. In the case of the KITTI dataset, our methods largely outperform all the competitors in all metrics. In terms of $\delta<1.25^2$ and $\delta<1.25^3$, we have only minor improvement since the results of these two metrics are nearly saturated. In the case of C+K, our results are significantly better than all the competitors except for the last two accuracy terms. We also found that our ResASPP model has a better performance than VGGASPP in all cases. 

\begin{table*}[!t]
\centering
\resizebox{\textwidth}{!}{\begin{tabular}{lccccccccccc}
    \hline
\multicolumn{1}{c}{\multirow{2}{*}{Approach}} &
\multicolumn{1}{c}{\multirow{2}{*}{Train}} & 
\multicolumn{1}{c}{\multirow{2}{*}{Test}} & 
\multicolumn{1}{c}{\multirow{2}{*}{PP}} & 
\multicolumn{1}{l}{ARD} & \multicolumn{1}{l}{SRD} & \multicolumn{1}{l}{RMSE} & \multicolumn{1}{l}{RMSE(log)}  &  & \multicolumn{1}{l}{$\delta<1.25$}  & \multicolumn{1}{l}{$\delta<1.25^2$}  & \multicolumn{1}{l}{$\delta<1.25^3$}  \\ 
\cline{5-8} \cline{10-12} \multicolumn{4}{c}{} & \multicolumn{4}{c}{Lower is better.} &  & \multicolumn{3}{c}{Higher is better.} \\ 
    \hline
Godard et al. \cite{Godard2017CVPR}	& K & E - 80m & Y & 0.1480 & 1.3440 & 5.927 & 0.247 & & 0.803 & 0.922 & 0.964 \\
Po-Yi et al. \cite{Poyi2019cvpr} - w/ seg & K & E -80m & Y & 0.1180 & 0.9050 & 5.096 & 0.211 & & 0.839 & \textbf{0.945} & \textbf{0.977} \\ 
%Ours VGGASPP & K & E - 80m & Y & 0.1086 & 0.9256 & 4.951 & 0.207 & & 0.857 & 0.942 & 0.974 \\
%Ours ResASPP & K & E - 80m & Y & 0.1089 & 0.9063 & 4.951 & 0.208 & & 0.857 & 0.941 & 0.974 \\
Ours VGGASPP & K & E - 80m & Y+ & \textbf{0.1070} & 0.9055 & 4.873 & \textbf{0.202} & & \textbf{0.862} & \textbf{0.945} & 0.975 \\
Ours ResASPP & K & E - 80m & Y+ & 0.1073 & \textbf{0.8849} & \textbf{4.866} & 0.203 & & \textbf{0.862} & \textbf{0.945} & 0.975 \\
    \hline
Godard et al. \cite{Godard2017CVPR}	& C+K & E - 80m & Y & 0.1140 & 0.8980 & 4.935 & 0.206 & & 0.861 & 0.949 & 0.976 \\
%Ours VGGASPP & C+K & E - 80m & Y & 0.1023 & 0.6634 & 3.725 & 0.194 & & 0.870 & 0.948 & 0.977 \\
%Ours ResASPP & C+K & E - 80m & Y & 0.1028 & 0.6602 & 3.724 & 0.195 & & 0.869 & 0.948 & 0.977 \\
Ours VGGASPP & C+K & E - 80m & Y+ & \textbf{0.1008} & 0.6462 & \textbf{3.653} & 0.190 & &  \textbf{0.875} & \textbf{0.952} & \textbf{0.979} \\
Ours ResASPP & C+K & E - 80m & Y+ & 0.1012 & \textbf{0.6425} & 3.645 & \textbf{0.190} & & \textbf{0.875} & \textbf{0.952} & \textbf{0.979} \\
    \hline
Godard et al. \cite{Godard2017CVPR}	& K & E - 50m & Y & 0.1400 & 0.9760 & 4.471 & 0.232 & & 0.818 & 0.931 & 0.969 \\
Po-Yi et al. \cite{Poyi2019cvpr} - w/ seg & K & E -50m & Y & 0.1120 & 0.6730 & 3.871 & 0.198 & & 0.852 & 0.951 & \textbf{0.980} \\
%Ours VGGASPP & K & E - 50m & Y & 0.1032 & 0.8155 & 4.732 & 0.198 & & 0.869 & 0.949 & 0.977 \\
%Ours ResASPP & K & E - 50m & Y & 0.1026 & 0.8282 & 4.746 & 0.199 & & 0.869 & 0.948 & 0.976 \\
Ours VGGASPP & K & E - 50m & Y+ & 0.1014 & \textbf{0.7942} & \textbf{4.627} & \textbf{0.192} & & \textbf{0.876} & \textbf{0.953} & 0.979 \\
Ours ResASPP & K & E - 50m & Y+ & \textbf{0.1008} & 0.8054 & 4.643 & 0.193 & & 0.875 & 0.952 & 0.979 \\
	\hline
Godard et al. \cite{Godard2017CVPR}	& C+K & E - 50m & Y & 0.1080 & 0.6570 & 3.729 & 0.194 & & 0.873 & 0.954 & 0.979 \\
%Ours VGGASPP & C+K & E - 50m & Y & 0.0977 & 0.6027 & 3.577 & 0.187 & & 0.881 & 0.954 & 0.979 \\
%Ours ResASPP & C+K & E - 50m & Y & 0.0968 & 0.6020 & 3.566 & 0.187 & & 0.880 & 0.954 & 0.979 \\
Ours VGGASPP & C+K & E - 50m & Y+ & 0.0959 & 0.5853 & 3.486 & \textbf{0.181} & & \textbf{0.887} & \textbf{0.958} & \textbf{0.981} \\
Ours ResASPP & C+K & E - 50m & Y+ & \textbf{0.0950} & \textbf{0.5847} & \textbf{3.474} & \textbf{0.181} & & \textbf{0.887} & \textbf{0.958} & \textbf{0.981} \\
    \hline
Godard et al. \cite{Godard2017CVPR}	& C+K & E - 80m(u) & Y & 0.1300 & 1.1970 & 5.222 & 0.226 & & 0.843 & 0.940 & 0.971 \\
%Ours VGGASPP	& C+K & E - 80m(u) & Y & 0.1156 & 1.0601 &  4.979 &  0.215 & &  0.855 &  0.942 &  0.973 \\ 
%Ours ResASPP	& C+K & E - 80m(u) & Y & 0.1147 & 1.0604 &  4.988 &  0.215 & &  0.855 &  0.942 &  0.973 \\
Ours VGGASPP	& C+K & E - 80m(u) & Y+ & 0.1141 & 1.0460 &  \textbf{4.884} &  \textbf{0.210} & &  \textbf{0.861} &  \textbf{0.946} &  \textbf{0.975} \\
Ours ResASPP	& C+K & E - 80m(u) & Y+ & \textbf{0.1132} & \textbf{1.0419} &  4.895 &  \textbf{0.210} & &  \textbf{0.861} &  \textbf{0.946} &  \textbf{0.975} \\
    \hline
\end{tabular}}
\caption{ This table shows the intgration benchmark specifically compared with compared with Godard et al. \cite{Godard2017CVPR} and Po-Yi et al. \cite{Poyi2019cvpr} with post-processing (PP). All the results still use the crop defined by Garg et al. \cite{Garg2016ECCV} except for the last section, in which we evaluate the uncropped(u) results. In the PP column, Y means conventional PP, while Y+ means the proposed Edge-Guided PP. Overall, our results are better than the reference method in any scenario. The second reference \cite{Poyi2019cvpr} uses semantic segmentation as a reference to enhance the performance. We show that our method outperforms \cite{Poyi2019cvpr} even without the information of semantic segmentation. An interesting finding is that the VGGASPP and ResASPP have less difference when the proposed Edge-Guided PP is applied.}
\label{tbl:lw-eg-benchmark-integration}
\end{table*}

In the qualitative benchmark of network test, we choose the best two priors \cite{Godard2017CVPR} and \cite{Alex2019cvpr} as our references in Fig. \ref{fig:demo-benchmark-network}. We demonstrate that our ResVGG has a better ability to capture either large size objects (e.g., trucks, cars, wall, etc.) and fine details (e.g., poles, trunks, signs, etc.). Our method has less false detection in the sky area.  Our results are more clear and consistent with either the ground truth or the input image compared to the priors. 

\begin{figure*}[t]
\centering
	\begin{subfigure}[t]{.18\textwidth}
		\centering
		\includegraphics[width=.45\textwidth]{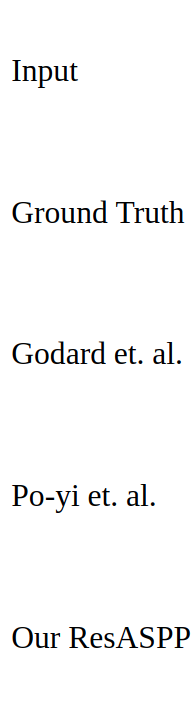}
	\end{subfigure}	
	\begin{subfigure}[t]{.2\textwidth}
		\centering
		\includegraphics[width=1.0\textwidth]{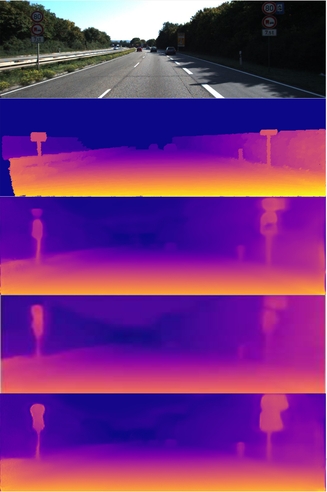}		
	\end{subfigure}
	\begin{subfigure}[t]{.2\textwidth}
		\centering
		\includegraphics[width=1.0\textwidth]{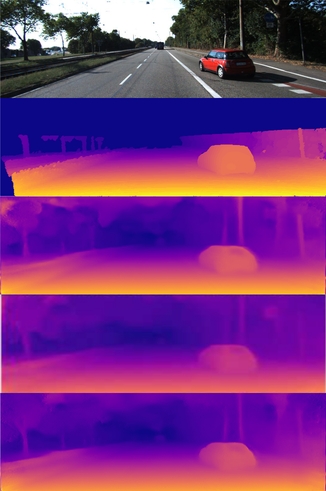} 
	\end{subfigure}
	\begin{subfigure}[t]{.2\textwidth}
		\centering
		\includegraphics[width=1.0\textwidth]{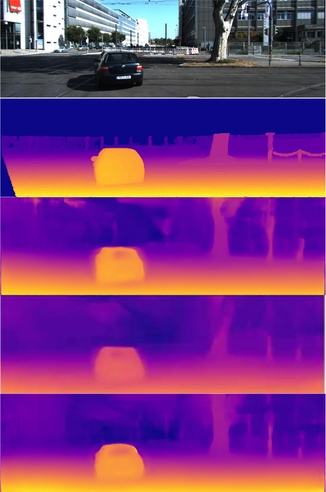}
	\end{subfigure}
	\begin{subfigure}[t]{.2\textwidth}
		\centering
		\includegraphics[width=1.0\textwidth]{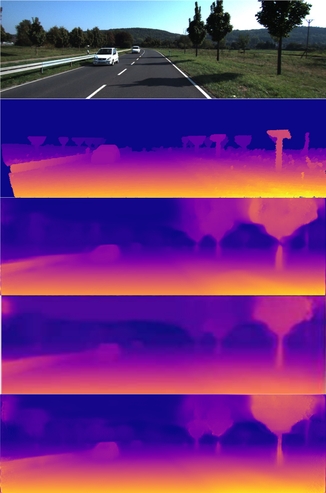}
	\end{subfigure}
\caption{Benchmark of qualitative result on KITTI dataset KITTI Split.
%We also interpolate the sparse ground truth velodyne depth for visualization.
There are eight sets in two parts listed in the Figure. The images from top to down in each set are - source image, ground truth, Godard et al. \cite{Godard2018arxiv}, Po-yi et. al. \cite{po2019cvpr}, and our ResASPP. 
%All the results have post-processing for a fair comparison. 
Our ResASPP has applied the proposed Edge-Guided post-processing. 
%We choose the same pictures to run the benchmark to compare with the method \cite{po2019cvpr} that considers the semantic segmentation consistency.
Our results can capture much more clear object shapes, such as signs, cars, and trunks than priors. The halo effects are also effectively reduced in our results. 
%In this comparison, we showcase that the proposed method can compete with the method that self-supervised by semantic segmentation. 
%$^\star$ The results of \cite{po2019cvpr} are cropped directly from source paper.
}
\label{fig:demo-benchmark-integration}
\end{figure*}

\paragraph{Integration Test} In the integration test of our benchmark, we involve the post-processing in the comparison. To further showcase the power of the propose Edge-Guided post-processing, we add a special prior \cite{po2019cvpr} that uses semantic consistency to capture better object shape in the estimated depth map. There are multiple scenarios in this test. From the training aspect of view, there are K only and C+K cases. In test cases, we still use Eigen-split with full and near distance under Garg \cite{Garg2016ECCV} crop, but an uncropped full distance comparison is added at the end of the quantitative results of Table \ref{tbl:lw-eg-benchmark-integration}. Our results with conventional post-processing are still better than two competitors except for the last two accuracy terms in all cases. When we apply the Edge-Guided post-processing method, our results become significantly better, and only the last accuracy term is slightly behind.   

The improved performance of our method is not only quantitative, but also qualitative. The results are shown in Fig. \ref{fig:demo-benchmark-integration}. We use the same demo pictures as \cite{po2019cvpr} to have a fair comparison. Our results have a much better ability to reproduce clear object shapes and edges in any size, especially the signs and trunks in the test images. The halo effects around objects (e.g., cars, signs, trunks,etc.) are largely reduced using the proposed Edge-Guided post-processing. In the visual evaluation, we provide much more accurate and visually appealing images to viewers.      

In the qualitative benchmark, we include more results on KITTI dataset and extend our model on more datasets. Please refer to Appendix to check more results.

\section{Discussion}

In our benchmark, we have proved that the proposed Light-weight DispNet is better than related methods. The only difference is the introduced ASPP module to replace the last few dense feature extractor in the encoder. ASPP module has a smaller structure than the conventional dense feature extractors but can better detect long-range features. 
%ASPP module has a minimal structure but has a better capability to detect long-range features than the conventional dense feature extractors. 

The quantitative results have shown the improvement of the ASPP module, but we still have some limitations in the last two accuracy terms. These two terms indicate the coarse accuracy of the estimation. We can find that we mainly improve the object shape and clarity with ASPP, but no significant improvement in the rest flat area or very large object. The only improvement of the large area in the sky area that is not counted in the evaluation because there is no ground truth in this area. The current design of ASPP module has no significant impact on these parts. The possible solution is to improve the ASPP module further, e.g., enlarge the detection range, or add local-global information \cite{JinarXiv2019} to align the rest non-object area. 

\begin{figure}[ht]
\centering
\begin{subfigure}{.22\textwidth}
  \centering
  \includegraphics[width=1.\linewidth]{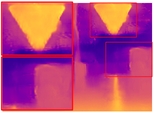}
  \caption{Conventional \cite{Godard2017CVPR}}
  \label{fig:sub1}
\end{subfigure}%
\begin{subfigure}{.22\textwidth}
  \centering
  \includegraphics[width=1.\linewidth]{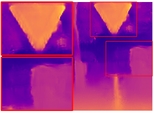}
  \caption{Ours Edge-Guided}
  \label{fig:sub2}
\end{subfigure}
\caption{Post-processing performance comparison}
\label{fig:lw-eg-limit}
\end{figure}

Although the halo effects have been largely removed using the Edge-Guided post-processing, there are still a few remaining halos. In this method, there are two assumptions: 1) the network can accurately capture the right-side edge of the object, and 2) the occlusion fading is inside the detection range. The first assumption indicates that the Edge-Guided method highly relies on the detection results of the network. If the network can only generate a blurred edge in the estimation, the performance of our method is limited. Furthermore, the detection range $N$ of our Edge-Guided method is also a trick. If we use a small $N$ in detection, the long occlusion fading can not be detected, and there will be a strong undershoot around the object contour in the final output. However, a large $N$ may involve unnecessary information to sabotage the detection results. Currently, we found that $N = 10$ is the most balance setting, although there is still a bit remaining long-range fading as shown in Fig. \ref{fig:lw-eg-limit}. It may need a better edge detector to overcome this limitation.
%This limitation may need a better edge detector to improve the performance.    

\vspace{-1em}

\section{Conclusion}

We have presented a Light-weight DispNet and a novel Edge-Guided post-processing method to improve the performance for a self-supervised monocular depth estimator. Our primary contribution is that the proposed Light-weight DispNet demonstrates the inherent capability to capture long-range features to better estimate the depth map with a much smaller network structure compared to the current commonly used DispNet. We get up to a $48\%$ speed up in inference and $71.8\%$ fewer parameters of the network with a better performance both quantitatively and qualitatively. Another contribution of this work is that the Edge-Guided post-processing can resolve most occlusion fading effect of self-supervision methods. It can effectively reduce the halo effect that comes from the conventional post-processing to yield the object shape. The proposed Edge-Guided post-processing is an upgraded version of the conventional post-processing \cite{Godard2017CVPR} and is suitable for all the self-supervised monocular depth estimators. The quantitative and qualitative results are both clear in the results. Our model only costs $4\%$ more in computation time, which is a very low price to improve the performance. 

% In the future, we would like to improve our network further to reach over 60 FPS and the performance can compete with supervision methods. In this paper, we only improve the encoder part of the DispNet and also not yet consider using MobileNet \cite{Andrew2017arXiv} structure. There is still some potential to reduce the network size further and improve the performance simultaneously. On the other hand, we are also seeking a better edge detector to enhance the Edge-Guided post-processing to reduce the occlusion fading and halo effects entirely. 

\section*{Acknowledgments}

This material is based upon work supported by the National Science Foundation under Grant Number 1622589 “Computer Guided Laparoscopy Training”. Any opinions, findings, and conclusions or recommendations expressed in this material are those of the authors and do not necessarily reflect the views of the National Science Foundation. G. Ditzler was supported by the Department of Energy under grant DE-NA0003946. 

\bibliographystyle{unsrt}  
\bibliography{ms}

\newpage

\appendix

\begin{appendices}

\begin{center}
\centering
{
\LARGE \textbf{Appendix}
}
\end{center}

\section{Network Architecture}

In this section, we provide the network designs of our methods. The VGG and Resnet$\_$blocks are as the same as the conventional designs, but we remove the batch norm process of Resnet$\_$block. The ASPP module is directly adopted from the prior \cite{Liang2017arxiv} in the implementation.

\begin{table}[ht]
    \begin{subtable}[t]{0.5\linewidth}
        \centering
        \resizebox{.9\textwidth}{!}{\begin{tabular}{ccccc} \hline
        Layers  & kernel & Ch I/O    & Scale    & input \\ \hline \hline 
        VGG\_block$\_$1 & 7    & 3/32      & 2  & left    \\ \hline
        VGG\_block$\_$2 & 5    & 32/64     & 4  & VGG\_block$\_$1    \\ \hline
        VGG\_block$\_$3 & 3    & 64/128    & 8  & VGG\_block$\_$2    \\ \hline
        VGG\_block$\_$4 & 3    & 128/256   & 16  & VGG\_block$\_$3   \\ \hline
        maxpool$\_$4 & 3       & 256/256   & 32  & VGG\_block$\_$4   \\ \hline
        ASPP    & - & 256/256    & 32   & maxpool$\_$4   \\ \hline
        \end{tabular}}
        \caption{Proposed VGGASPP encoder}
        \label{tbl:lw-eg-vggaspp}
    \end{subtable}
    \begin{subtable}[t]{0.5\linewidth}
        \centering
        \resizebox{.9\textwidth}{!}{\begin{tabular}{ccccc} \hline
        Layers  & kernel & Ch I/O    & Scale    & input \\ \hline \hline 
        enc\_block\_1 & conv$\_$7 & 7      & 2/64	    & 2     \\ \hline
        enc\_block\_2 & maxpool$\_$3 & 3	&64/64	    & 4     \\ \hline
        enc\_block\_3 & Resnet\_block & 3  & 64/64     & 8     \\ \hline
        enc\_block\_4 & Resnet\_block & 4  & 64/128    & 16    \\ \hline
        enc\_block\_5 & maxpool$\_$3 & 3   & 128/128   & 32    \\ \hline
        enc\_block\_6 & ASPP   & 128/256    &32 \\ \hline
        \end{tabular}}
        \caption{Proposed ResASPP encoder}
        \label{tbl:lw-eg-resaspp}
    \end{subtable}   \\
    \begin{subtable}[t]{0.5\linewidth}
        \centering
        \begin{tabular}{ccccc}
        \hline
        \multicolumn{5}{c}{VGG$\_$block} \\ \hline\hline
        layers  & kernel   & stride    & scale   & input \\ \hline
        conv1  & k & 2 & 2 & block$\_$input \\ \hline
        conv2  & k & 1 & 1 & conv1 \\ \hline
        \multicolumn{5}{c}{Resnet$\_$block} \\ \hline\hline
        layers  & kernel   & stride    & scale   & input \\ \hline
        conv1  & 1 & 1 & 1 & block$\_$input \\ \hline
        conv2  & 3 & 2 & 2 & conv1 \\ \hline
        conv3  & 1 & 1 & 1 & conv2 \\ \hline
        proj    & - & - & - & \begin{tabular}[c]{c}block$\_$input+\\ conv3\end{tabular}
        \end{tabular}
        \caption{Backbone basic blocks}
        \label{tab:basic_block}
    \end{subtable}
    \begin{subtable}[t]{0.5\linewidth}
        \centering    
        \begin{tabular}{ccccc}
        \hline
        \multicolumn{5}{c}{ASPP} \\ \hline
        layers  & kernel   & rate    & Ch O   & input \\ \hline
        reduce$\_$mean  & - & -     & no change & block$\_$input \\ \hline
        conv1   & 1 & - & no change & reduce$\_$mean \\ \hline
        upscale   & - & - & no change & conv1 \\ \hline
        Astrous1    & 1 & 1 & 256 & block$\_$input \\ \hline
        Astrous6    & 3 & 6 & 256 & block$\_$input \\ \hline
        Astrous12   & 3 & 12 & 256 & block$\_$input \\ \hline
        Astrous18   & 3 & 18 & 256 & block$\_$input \\ \hline 
        concat   & - & - & - & 
        \begin{tabular}[c]{c}upscale+\\ Astrous1+\\ Astrous6+\\ Astrous12+\\ Astrous18 \end{tabular}
        \\ \hline 
        \end{tabular}
    \caption{ASPP Module}
    \label{tab:ASPP_block}        
    \end{subtable}     
    \caption{Proposed Network architecture, where k is the kernel size, Ch I/O is the number of input and output channels for each layer, Scale is the down-scaling factor for each layer relative to the input image, and input corresponds to the input of each layer where + is a concatenation and $\ast$ is a 2$\times$ upsampling of the layer.}
    \label{tab:lw-eg-network-details}        
\end{table}

\section{Edge-Guided Post-Processing}
One of our major contribution is the Edge-Guided post-processing method that evolves from \cite{Godard2017CVPR}. We further elaborate some detailed implementations that we haven't mentioned in the main manuscript. 

The conventional post-processing of \cite{Godard2017CVPR} can be formulated as follows:
\begin{equation}
\begin{aligned}
&d_{l-syn} = d_lw_l + d''_lw'_l + d_{lm}(1-w_l-w'_l) \\
&d_{lm} = (d_l + d''_l)/2 \\
&w_l(i,j) = 
\begin{cases}
    1&\text{if $j\leq{rng}$} \\
    0&\text{if $j>{rng}$} \\
    1-20\times(rng-j)&\text{else} \\
\end{cases}
\end{aligned}
\label{eq:ss-sup-depth-loss}
\end{equation}
where $d_{l-syn}$ is the synthesized left disparity, $d_{lm}$ is the modulated left disparity by $d_l$ and $d''_l$, $(i,j)$ are normalized pixel coordinates, $w_l$ is the per-pixel weight map, and $rng$ is the reserved boundary range that was set as 0.05 in \cite{Godard2017CVPR}. This equation set shows that the left and right side has $5\%$ reserved range for $d_l$ and $d''_l$, while the center part uses $d_{lm}$ that is the the average of $d_l$ and $d''_l$. This is a boundary preserving design to avoid the boundary fading during the disparity synthesis in the post processing. 

In our Edge-Guided post-processing, we introduced the edge-weighted left disparity as shown in Eq. (5) of the main script:
\begin{equation}
d^{\star}_l = wd_l+w''d_l
\end{equation}
We also follow the boundary reserving design here and set the modulated left disparity $d_{lm}$ as $d^{\star}_l$. Furthermore, the reserved range $rng$ is only $2\%$ in our method. The final Edge-guided synthesized left disparity is as following:
\begin{equation}
d^{\star}_{l-syn} = d_lw_l + d''_lw'_l + d^{\star}_l(1-w_l-w'_l) \\
\end{equation}

\begin{figure}[!ht]
\begin{center}
	\includegraphics[width=0.95\textwidth]{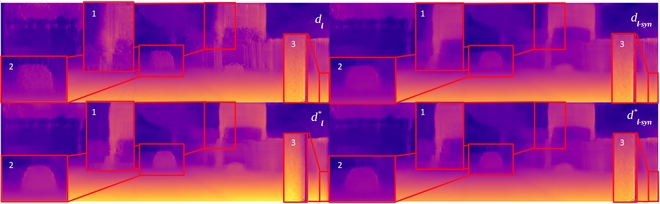}
\end{center}	
\caption{Comparison between the conventional post-processing (PP) \cite{Godard2017CVPR} and the proposed Edge-Guided post-processing (EG-PP) on KITTI dataset.
$d_l$ is the left disparity, $d_{l-syn}$ is the synthesized left disparity of \cite{Godard2017CVPR}, $d^{\star}_{l}$ is the proposed edge-weighted left disparity without the boundary preserving, and $d^{\star}_{l-syn}$ is the proposed edge-weighted left disparity with the boundary preserving. Our $d^{\star}_{l}$ and $d^{\star}_{l-syn}$ can reduce the occlusion fading well with minimal halo effects compared to $d_{l-syn}$ shown in the bounding boxes 1 and 2. The boundary preserving method works effectively to reduce the boundary gradation shown in the bounding boxes 3 of $d^{\star}_{l}$. }
\label{fig:lw-eg-sup-ed-explain}
\end{figure}

We visualize the results of $d_l$, $d_{l-syn}$, $d^{\star}_{l}$, and $d^{\star}_{l-syn}$ to show the improvement of our method in Fig. \ref{fig:lw-eg-sup-ed-explain}. In the bounding boxes 1 and 2 show the occlusion fading reduction with the minimal halo effect of our method. In the bounding box 3, the boundary gradation of $d^{\star}_{l}$ can be improved by using the boundary preserving method. The $d^{\star}_{l}$ and $d^{\star}_{l-syn}$ in Fig. \ref{fig:lw-eg-sup-ed-explain} have a few color difference of the color-mapping due to the value range variation between two results.  

\section{Experiments}
There are two updates in our experiments setup. The first update is the training setup. We originally used 8 batches and 100 epochs (8x100) on the KITTI training dataset, while in the C+K training case we train (8x50) on Cityscapes dataset and another 8x100 on KITTI dataset. However, we actually only used 8x50 on KITTI dataset of the C+K training case to get a optimal result and we would like to correct this wrong description. 

The second update is the parameter optimization of the reserved boundary range ($rng$) in the Edge-Guiding post-processing. Originally, we applied the same $rng$ (0.05) as the prior \cite{Godard2017CVPR} and we further optimized it to 0.02 due the narrower boundary gradation in our method compared to prior.

Based on these two updates, we also update out experimental results of the ablation study and the benchmarks. We also include more comprehensive comparison of our method with the competitors. 

\subsection{Ablation Study}

We have updated the results of the Ablation Study in Table \ref{tbl:sup-ablation-nums}. The computational costs have no change in our new setup compared to the original one, so there is no update here. The performance of the proposed method has a bit improvement under the new setup and the tendency of the results keep the same. In the KITTI slpit, the proposed VGGASPP and ResASPP has similar performance on KITTI training case. The ResASPP model has overall better performance than the VGGASPP model on KITTI training case. However, the situation is opposite on the C+K training case. The VGGASPP model has a better results than ResASPP. 

In Table \ref{tbl:sup-ablation-nums}, we would also like to update the results of our VGGASPP on the KITTI training case. We used a incomplete training model to produce the results in the main manuscript. Here we provide the correct results that have better performance.

\begin{table*}[!ht]
\centering
\resizebox{\textwidth}{!}{\begin{tabular}{lccccccccccc}
    \hline
\multicolumn{1}{c}{\multirow{2}{*}{Approach}} & 
\multicolumn{1}{c}{\multirow{2}{*}{Train}} & 
\multicolumn{1}{c}{\multirow{2}{*}{PP}} & 
\multicolumn{1}{l}{ARD} & \multicolumn{1}{l}{SRD} & \multicolumn{1}{l}{RMSE} & \multicolumn{1}{l}{RMSE(log)}  & \multicolumn{1}{l}{D1-all}  &  & \multicolumn{1}{l}{$\delta<1.25$}  & \multicolumn{1}{l}{$\delta<1.25^2$}  & \multicolumn{1}{l}{$\delta<1.25^3$} \\ 
\cline{4-8} \cline{10-12} \multicolumn{3}{c}{} & \multicolumn{5}{c}{Lower is better.} &  & \multicolumn{3}{c}{Higher is better.} \\  
    \hline
Godard et al. \cite{Godard2017CVPR} VGG & K & N & 0.1240 & 1.3880 & 6.125 & 0.217 & 30.272 & & 0.841 & 0.936 & 0.975 \\
Godard et al. \cite{Godard2017CVPR} Resnet50 & K & N & 0.1127 & 1.1331 & 5.749 & 0.203 & 29.553 & & 0.851 & 0.947 & 0.980 \\
    \hline
Our VGGASPP & K & N & 0.1134 & 1.1636 & 5.734 & 0.201 & 27.379 & & 0.853 & 0.945 & 0.979 \\
Our VGGASPP & K & Y & 0.1079 & 1.0259 & 5.464 & 0.192 & 26.395 & & 0.857 & 0.949 & 0.982 \\
Our VGGASPP & K & Y+ & 0.1077 & 1.0238 & 5.387 & 0.189 & 26.152 & & 0.860 & 0.951 & 0.983 \\
Our ResASPP & K & N & 0.1107 & 1.0633 & 5.612 & 0.199 & 27.531 & & 0.854 & 0.946 & 0.980 \\
Our ResASPP & K & Y & 0.1075 & 0.9878 & 5.474 & 0.193 & 27.050 & & 0.855 & 0.949 & 0.981 \\
Our ResASPP & K & Y+ & 0.1071 & 0.9936 & 5.394 & 0.189 & 26.673 & & 0.858 & 0.952 & 0.983 \\
    \hline
\iffalse
Godard et al. \cite{Godard2017CVPR} VGG & C+K & N 
 & 0.104 & 1.070 & 5.417 & 0.188 & 25.523 & & 0.875 & 0.956 & 0.983 \\
Godard et al. \cite{Godard2017CVPR} VGG & C+K & Y & 0.100 & 0.934 & 5.141 & 0.178 & 25.077 & & 0.878 & 0.961 & 0.986 \\
Godard et al. \cite{Godard2017CVPR} Resnet50 & C+K & Y & \textbf{0.097} & 
\textbf{0.896} & 5.093 & 0.176 & 23.811 & & 0.879 & \textbf{0.962} & 0.986 \\ 
Our VGGASPP & C+K & Y & 0.0987 & 0.9130 & 5.129 & 0.179 & 23.361 & & 0.879 & 0.958 & 0.985 Our ResASPP & C+K & Y & 0.1003 & 0.9671 & 5.160 & 0.179 & 23.633 & & 0.875 & 0.958 & 0.986 \\
\\
\fi
Our VGGASPP & C+K & Y+ & \textbf{0.0984} & \textbf{0.9196} & \textbf{5.035} & \textbf{0.175} & \textbf{22.942} & & \textbf{0.883} & \textbf{0.961} & 0.986 \\
Our ResASPP & C+K & Y+ & 0.1000 & 0.9697 & 5.070 & 0.175 & 23.231 & & 0.879 & \textbf{0.961} & \textbf{0.987} \\
    \hline
\end{tabular}}
\caption{Quantitative results for different variants of our approach on the KITTI Stereo 2015 test dataset. The PP means using post-processing. N means no PP, Y means conventional PP, and Y+ represents the proposed Edge-Guided PP. The best result in each subsection is shown in bold. The training scenario is based on the KITTI training set (K), while the last section shows the results which are pre-trained by Cityscapes training sets (C+K). We use our prior \cite{Godard2017CVPR} as our baseline is shown in the first section. 
} 
\label{tbl:sup-ablation-nums}
\end{table*}

\subsection{Benchmarks}
The benchmark results are also updated base on the new setup. 
The network test has no update here, while the integration test is being updated due to post processing setup update and shown in Table \ref{tbl:lw-eg-sup-benchmark-integration}. We also add the results of our methods with conventional post-processing to indicate the improvement by using the proposed Edge-Guided method. The conclusion keeps consistent as the main manuscript and the performance is slightly improved under the new setup. 

\begin{table*}[!ht]
\centering
\resizebox{\textwidth}{!}{\begin{tabular}{lccccccccccc}
    \hline
\multicolumn{1}{c}{\multirow{2}{*}{Approach}} &
\multicolumn{1}{c}{\multirow{2}{*}{Train}} & 
\multicolumn{1}{c}{\multirow{2}{*}{Test}} & 
\multicolumn{1}{c}{\multirow{2}{*}{PP}} & 
\multicolumn{1}{l}{ARD} & \multicolumn{1}{l}{SRD} & \multicolumn{1}{l}{RMSE} & \multicolumn{1}{l}{RMSE(log)}  &  & \multicolumn{1}{l}{$\delta<1.25$}  & \multicolumn{1}{l}{$\delta<1.25^2$}  & \multicolumn{1}{l}{$\delta<1.25^3$}  \\ 
\cline{5-8} \cline{10-12} \multicolumn{4}{c}{} & \multicolumn{4}{c}{Lower is better.} &  & \multicolumn{3}{c}{Higher is better.} \\ 
    \hline
Godard et al. \cite{Godard2017CVPR}	& K & E - 80m & Y & 0.1480 & 1.3440 & 5.927 & 0.247 & & 0.803 & 0.922 & 0.964 \\
Po-Yi et al. \cite{Poyi2019cvpr} - w/ seg & K & E -80m & Y & 0.1180 & 0.9050 & 5.096 & 0.211 & & 0.839 & \textbf{0.945} & \textbf{0.977} \\ 
Ours VGGASPP & K & E - 80m & Y & 0.1086 & 0.9256 & 4.951 & 0.207 & & 0.857 & 0.942 & 0.974 \\
Ours ResASPP & K & E - 80m & Y & 0.1089 & 0.9063 & 4.951 & 0.208 & & 0.857 & 0.941 & 0.974 \\
Ours VGGASPP & K & E - 80m & Y+ & \textbf{0.1070} & 0.9055 & 4.873 & \textbf{0.202} & & \textbf{0.862} & \textbf{0.945} & 0.975 \\
Ours ResASPP & K & E - 80m & Y+ & 0.1073 & \textbf{0.8849} & \textbf{4.866} & 0.203 & & \textbf{0.862} & \textbf{0.945} & 0.975 \\
    \hline
Godard et al. \cite{Godard2017CVPR}	& C+K & E - 80m & Y & 0.1140 & 0.8980 & 4.935 & 0.206 & & 0.861 & 0.949 & 0.976 \\
Ours VGGASPP & C+K & E - 80m & Y & 0.1023 & 0.6634 & 3.725 & 0.194 & & 0.870 & 0.948 & 0.977 \\
Ours ResASPP & C+K & E - 80m & Y & 0.1028 & 0.6602 & 3.724 & 0.195 & & 0.869 & 0.948 & 0.977 \\
Ours VGGASPP & C+K & E - 80m & Y+ & \textbf{0.1008} & 0.6462 & \textbf{3.653} & 0.190 & &  \textbf{0.875} & \textbf{0.952} & \textbf{0.979} \\
Ours ResASPP & C+K & E - 80m & Y+ & 0.1012 & \textbf{0.6425} & 3.645 & \textbf{0.190} & & \textbf{0.875} & \textbf{0.952} & \textbf{0.979} \\
    \hline
Godard et al. \cite{Godard2017CVPR}	& K & E - 50m & Y & 0.1400 & 0.9760 & 4.471 & 0.232 & & 0.818 & 0.931 & 0.969 \\
Po-Yi et al. \cite{Poyi2019cvpr} - w/ seg & K & E -50m & Y & 0.1120 & 0.6730 & 3.871 & 0.198 & & 0.852 & 0.951 & \textbf{0.980} \\
Ours VGGASPP & K & E - 50m & Y & 0.1032 & 0.8155 & 4.732 & 0.198 & & 0.869 & 0.949 & 0.977 \\
Ours ResASPP & K & E - 50m & Y & 0.1026 & 0.8282 & 4.746 & 0.199 & & 0.869 & 0.948 & 0.976 \\
Ours VGGASPP & K & E - 50m & Y+ & 0.1014 & \textbf{0.7942} & \textbf{4.627} & \textbf{0.192} & & \textbf{0.876} & \textbf{0.953} & 0.979 \\
Ours ResASPP & K & E - 50m & Y+ & \textbf{0.1008} & 0.8054 & 4.643 & 0.193 & & 0.875 & 0.952 & 0.979 \\
	\hline
Godard et al. \cite{Godard2017CVPR}	& C+K & E - 50m & Y & 0.1080 & 0.6570 & 3.729 & 0.194 & & 0.873 & 0.954 & 0.979 \\
Ours VGGASPP & C+K & E - 50m & Y & 0.0977 & 0.6027 & 3.577 & 0.187 & & 0.881 & 0.954 & 0.979 \\
Ours ResASPP & C+K & E - 50m & Y & 0.0968 & 0.6020 & 3.566 & 0.187 & & 0.880 & 0.954 & 0.979 \\
Ours VGGASPP & C+K & E - 50m & Y+ & 0.0959 & 0.5853 & 3.486 & \textbf{0.181} & & \textbf{0.887} & \textbf{0.958} & \textbf{0.981} \\
Ours ResASPP & C+K & E - 50m & Y+ & \textbf{0.0950} & \textbf{0.5847} & \textbf{3.474} & \textbf{0.181} & & \textbf{0.887} & \textbf{0.958} & \textbf{0.981} \\
    \hline
Godard et al. \cite{Godard2017CVPR}	& C+K & E - 80m(u) & Y & 0.1300 & 1.1970 & 5.222 & 0.226 & & 0.843 & 0.940 & 0.971 \\
Ours VGGASPP	& C+K & E - 80m(u) & Y & 0.1156 & 1.0601 &  4.979 &  0.215 & &  0.855 &  0.942 &  0.973 \\ 
Ours ResASPP	& C+K & E - 80m(u) & Y & 0.1147 & 1.0604 &  4.988 &  0.215 & &  0.855 &  0.942 &  0.973 \\
Ours VGGASPP	& C+K & E - 80m(u) & Y+ & 0.1141 & 1.0460 &  \textbf{4.884} &  \textbf{0.210} & &  \textbf{0.861} &  \textbf{0.946} &  \textbf{0.975} \\
Ours ResASPP	& C+K & E - 80m(u) & Y+ & \textbf{0.1132} & \textbf{1.0419} &  4.895 &  \textbf{0.210} & &  \textbf{0.861} &  \textbf{0.946} &  \textbf{0.975} \\
    \hline
\end{tabular}}
\caption{ This table shows the additional benchmark specifically compared with compared with Godard et al. \cite{Godard2017CVPR} and Po-Yi et al. \cite{Poyi2019cvpr} with post-processing (PP). All the results still use the crop defined by Garg et al. \cite{Garg2016ECCV} except for the last section, in which we evaluate the uncropped(u) results. In the PP column, Y means conventional PP, while Y+ means the proposed Edge-Guided PP. Overall, our results are better than the reference method in any scenario. The second reference \cite{Poyi2019cvpr} uses semantic segmentation as a reference to enhance the performance. We show that our method outperforms \cite{Poyi2019cvpr} even without the information of semantic segmentation. An interesting finding is that the VGGASPP and ResASPP have less difference when the proposed Edge-Guided PP is applied.}
\label{tbl:lw-eg-sup-benchmark-integration}
\end{table*}

\section{More Quantitative Results}
\begin{figure*}[!ht]
\centering
	\begin{subfigure}[t]{.19\textwidth}
		\centering
		\includegraphics[width=1.\textwidth]{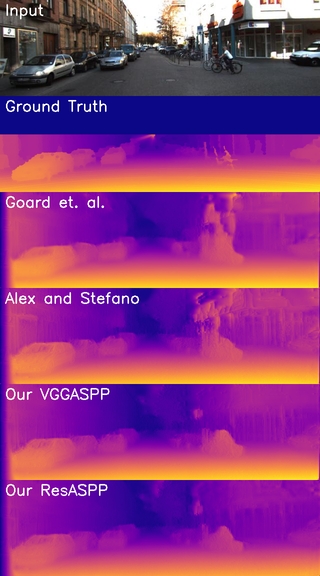}
	\end{subfigure}	
	\begin{subfigure}[t]{.19\textwidth}
		\centering
		\includegraphics[width=1.\textwidth]{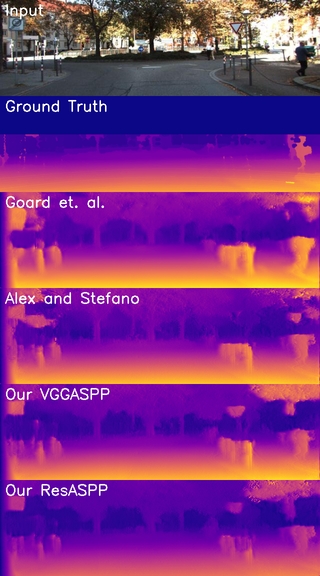}		
	\end{subfigure}
	\begin{subfigure}[t]{.19\textwidth}
		\centering
		\includegraphics[width=1.\textwidth]{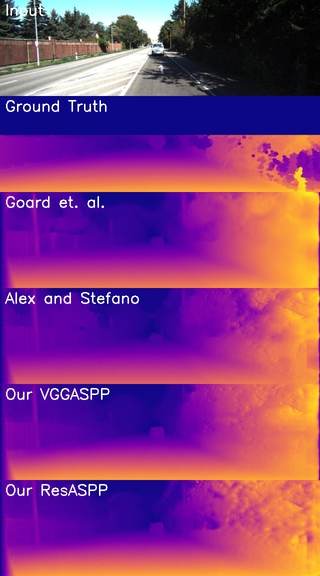} 
	\end{subfigure}
	\begin{subfigure}[t]{.19\textwidth}
		\centering
		\includegraphics[width=1.\textwidth]{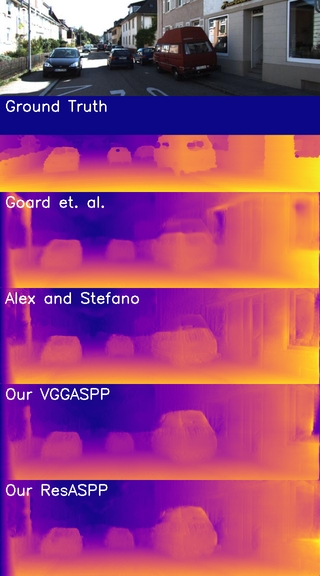}
	\end{subfigure}
	\begin{subfigure}[t]{.19\textwidth}
		\centering
		\includegraphics[width=1.\textwidth]{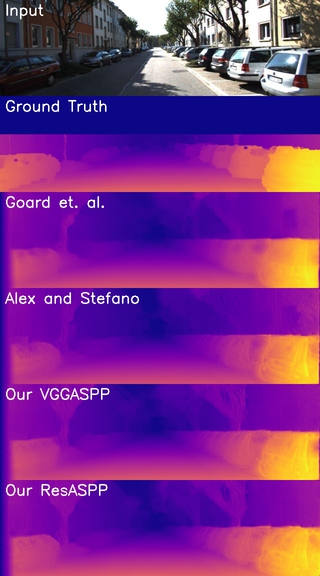}
	\end{subfigure} 
\caption{More More qualitative results of network benchmark}
\label{fig:lw-eg-sup-demo-benchmark-network}
\end{figure*}

We include more quantitative results to further visualize the performance of our model on KITTI dataset. The proposed VGGASPP and ResASPP networks are both presented in all the results. In the network benchmark shown in Fig. \ref{fig:lw-eg-sup-demo-benchmark-network}, we use our prior \cite{Godard2017CVPR} and \cite{Alex2019cvpr} as the references without any post-processing to demonstrate the network performance. The performance is the same as that we have shown in the main manuscript. Our network can detect more objects clearly than state-of-the-art models, such as signs, poles, cars, etc. Comparing to our VGGASPP and ResASPP, the overall results are similar, but our ResASPP can get a few better prediction results on object shapes. 

\begin{figure*}[ht]
\centering
	\begin{subfigure}[t]{.19\textwidth}
		\centering
		\includegraphics[width=1.\textwidth]{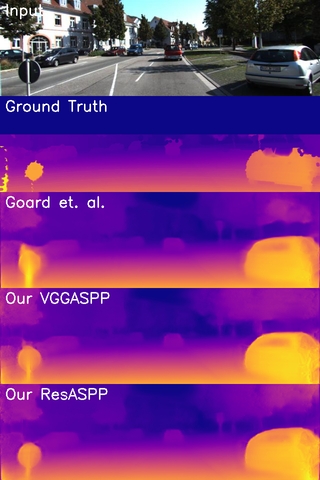}
	\end{subfigure}	
	\begin{subfigure}[t]{.19\textwidth}
		\centering
		\includegraphics[width=1.\textwidth]{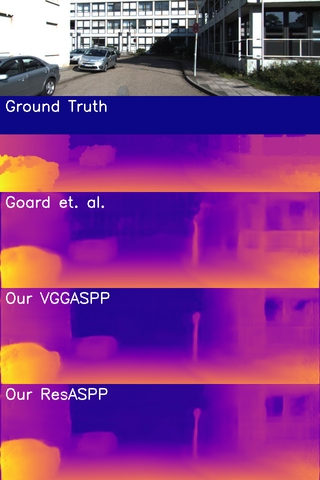}		
	\end{subfigure}
	\begin{subfigure}[t]{.19\textwidth}
		\centering
		\includegraphics[width=1.\textwidth]{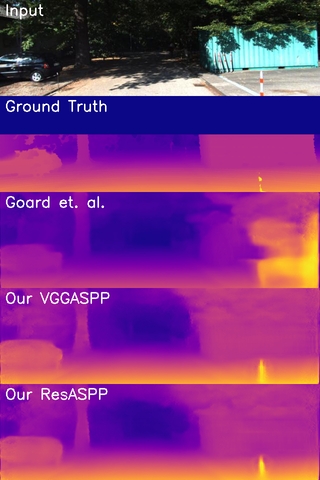} 
	\end{subfigure}
	\begin{subfigure}[t]{.19\textwidth}
		\centering
		\includegraphics[width=1.\textwidth]{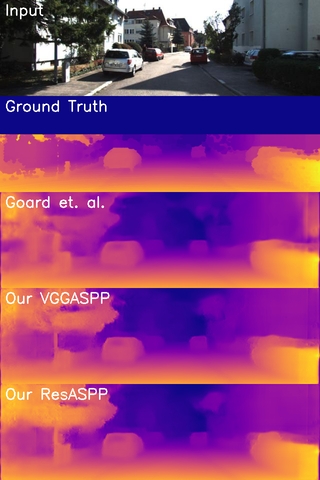}
	\end{subfigure}
	\begin{subfigure}[t]{.19\textwidth}
		\centering
		\includegraphics[width=1.\textwidth]{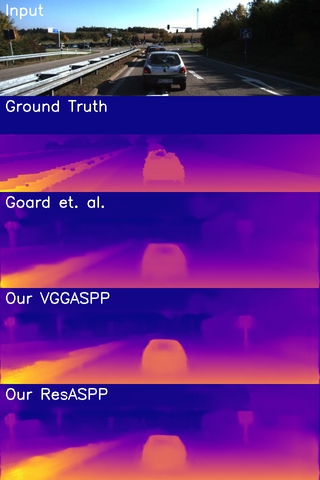}
	\end{subfigure} 
\caption{More qualitative results of Integration benchmark}
\label{fig:lw-eg-sup-demo-benchmark-integration}
\end{figure*}

In the integration benchmark shown in Fig. \ref{fig:lw-eg-sup-demo-benchmark-integration}, we only have one prior \cite{Godard2017CVPR} as the reference due to the lack of the source information from other priors. We further showcase the ability to capture the clear object shapes with the minimal halo effects in our methods. We can produce very clear disparity images in the inference using the proposed Edge-Guided post-processing method. 

\section{Extended Results On More Datasets }

\begin{figure*}[!ht]
\centering
	\begin{subfigure}[t]{.19\textwidth}
		\centering
		\includegraphics[width=1.\textwidth]{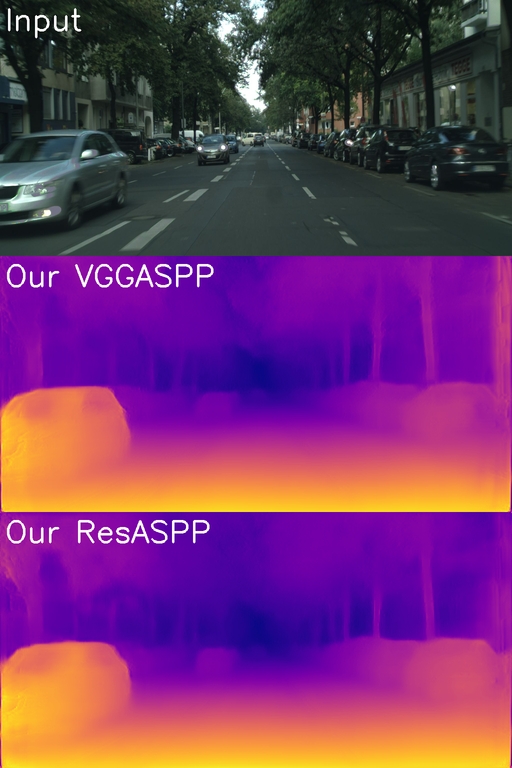}
	\end{subfigure}	
	\begin{subfigure}[t]{.19\textwidth}
		\centering
		\includegraphics[width=1.\textwidth]{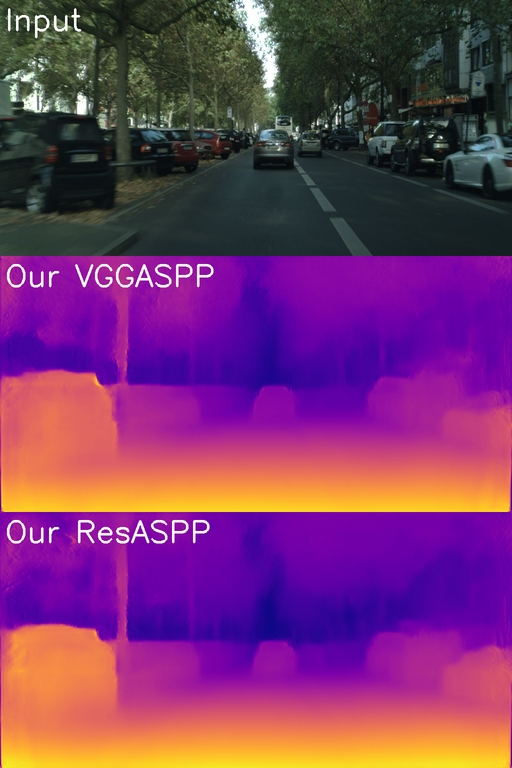}		
	\end{subfigure}
	\begin{subfigure}[t]{.19\textwidth}
		\centering
		\includegraphics[width=1.\textwidth]{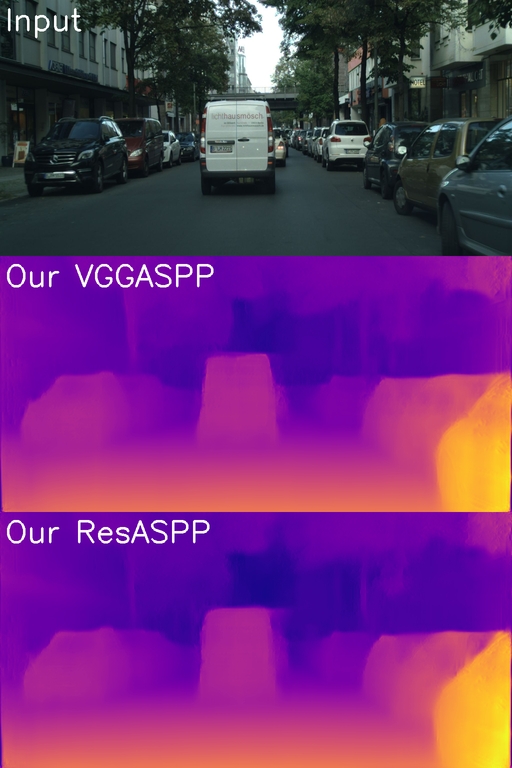} 
	\end{subfigure}
	\begin{subfigure}[t]{.19\textwidth}
		\centering
		\includegraphics[width=1.\textwidth]{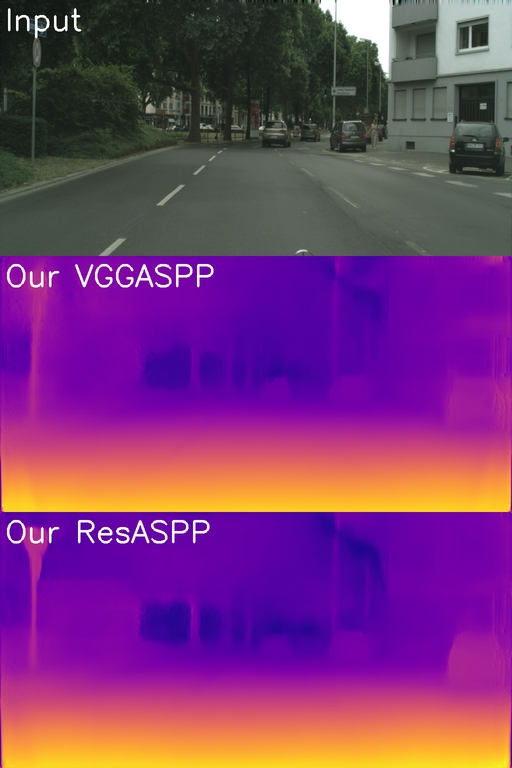}
	\end{subfigure}
	\begin{subfigure}[t]{.19\textwidth}
		\centering
		\includegraphics[width=1.\textwidth]{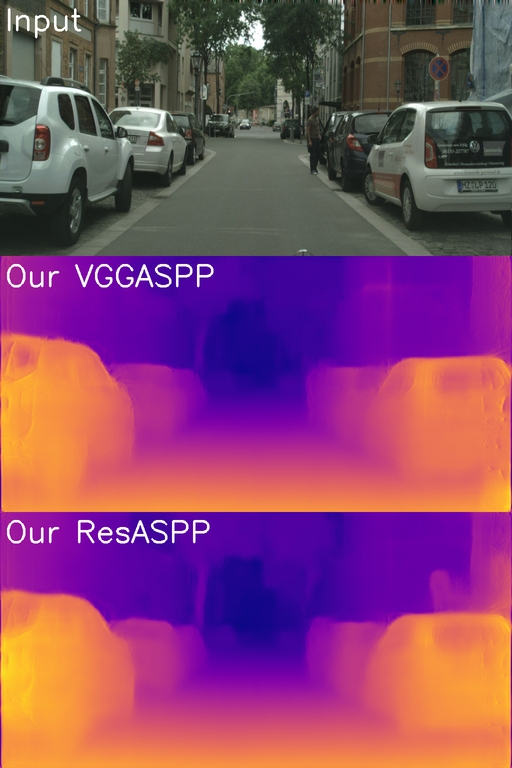}
	\end{subfigure} \\
	\begin{subfigure}[t]{.19\textwidth}
		\centering
		\includegraphics[width=1.\textwidth]{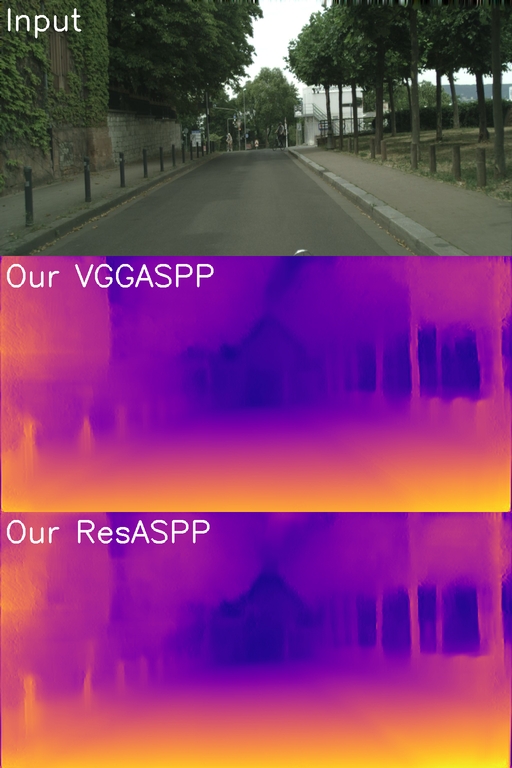}
	\end{subfigure}	
	\begin{subfigure}[t]{.19\textwidth}
		\centering
		\includegraphics[width=1.\textwidth]{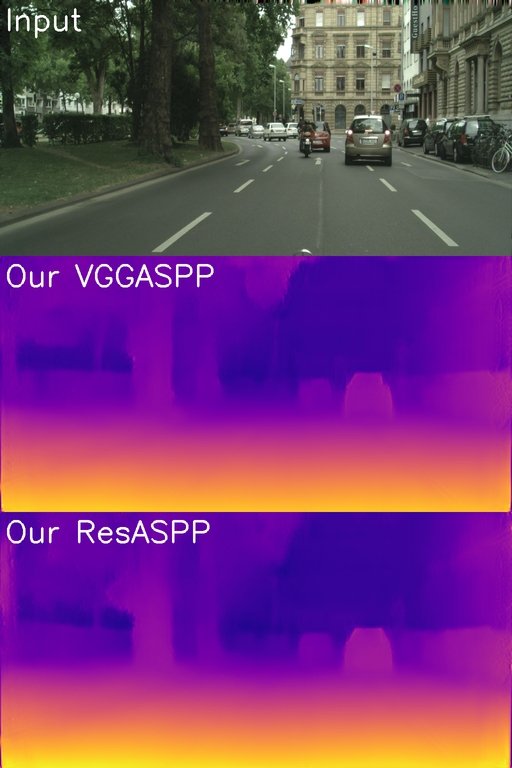}		
	\end{subfigure}
	\begin{subfigure}[t]{.19\textwidth}
		\centering
		\includegraphics[width=1.\textwidth]{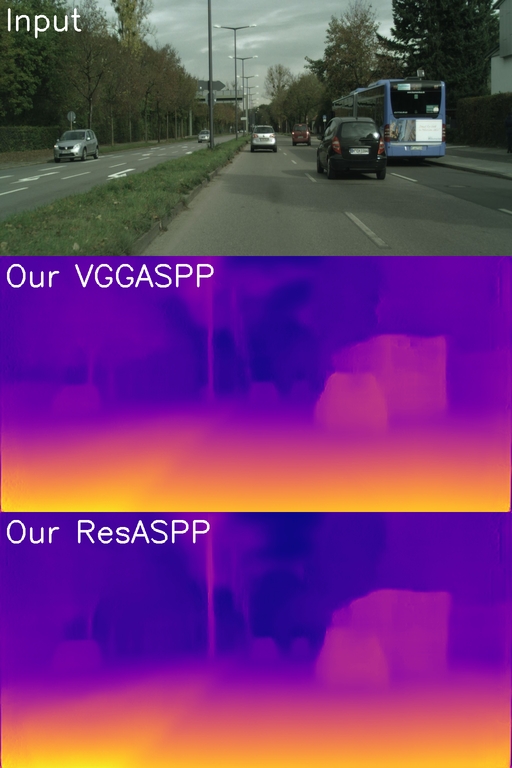} 
	\end{subfigure}
	\begin{subfigure}[t]{.19\textwidth}
		\centering
		\includegraphics[width=1.\textwidth]{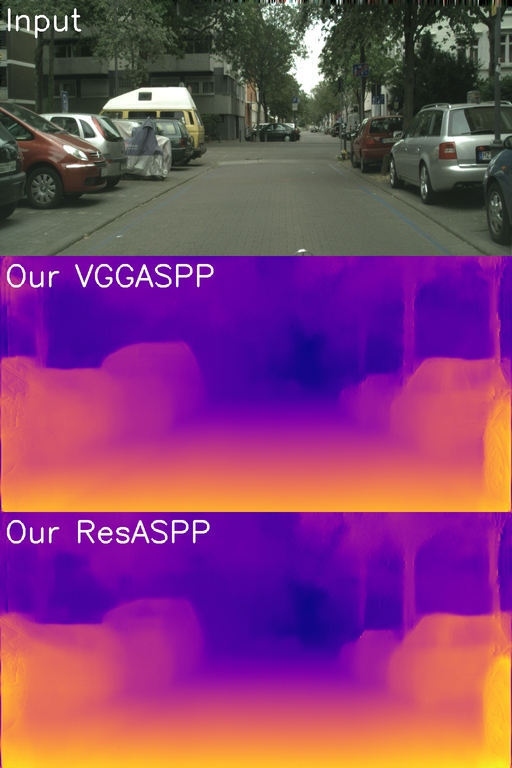}
	\end{subfigure}
	\begin{subfigure}[t]{.19\textwidth}
		\centering
		\includegraphics[width=1.\textwidth]{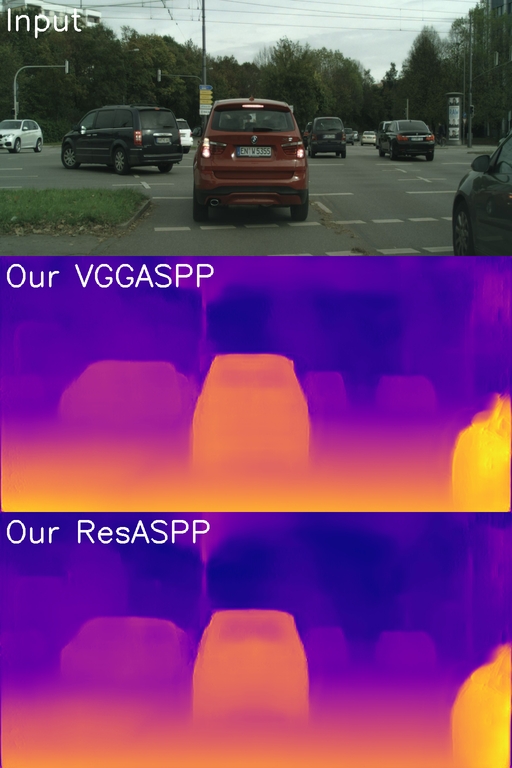}
	\end{subfigure} 
\caption{Cityscapes dataset qualitative results. There are 10 sets of images that we randomly select to show the visualization results.}
\label{fig:lw-eg-sup-demo-cityscapes}
\end{figure*}

\begin{figure*}[!ht]
\centering
	\begin{subfigure}[t]{.19\textwidth}
		\centering
		\includegraphics[width=1.\textwidth]{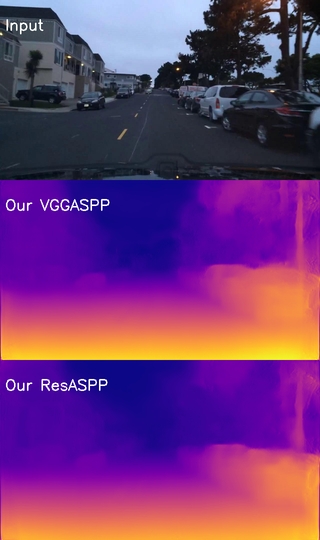}
	\end{subfigure}	
	\begin{subfigure}[t]{.19\textwidth}
		\centering
		\includegraphics[width=1.\textwidth]{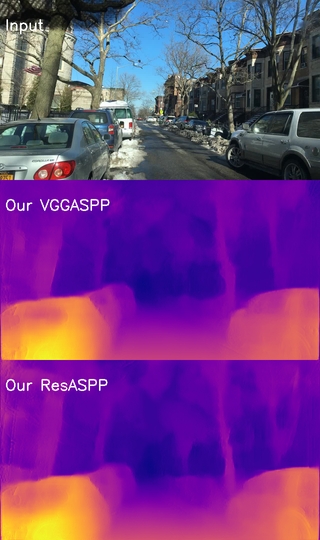}		
	\end{subfigure}
	\begin{subfigure}[t]{.19\textwidth}
		\centering
		\includegraphics[width=1.\textwidth]{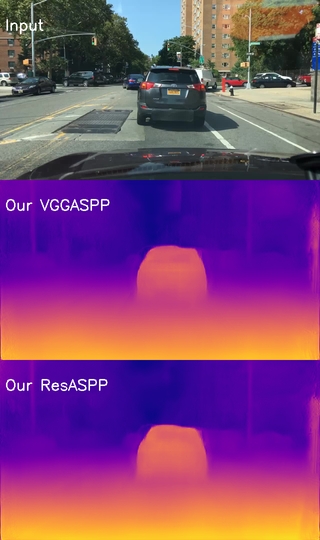} 
	\end{subfigure}
	\begin{subfigure}[t]{.19\textwidth}
		\centering
		\includegraphics[width=1.\textwidth]{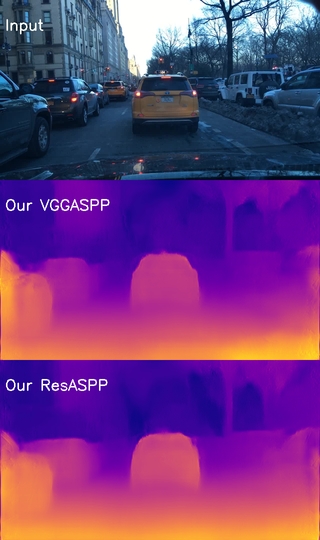}
	\end{subfigure}
	\begin{subfigure}[t]{.19\textwidth}
		\centering
		\includegraphics[width=1.\textwidth]{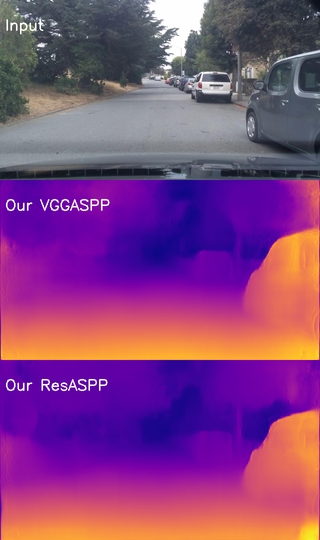}
	\end{subfigure} \\
	\begin{subfigure}[t]{.19\textwidth}
		\centering
		\includegraphics[width=1.\textwidth]{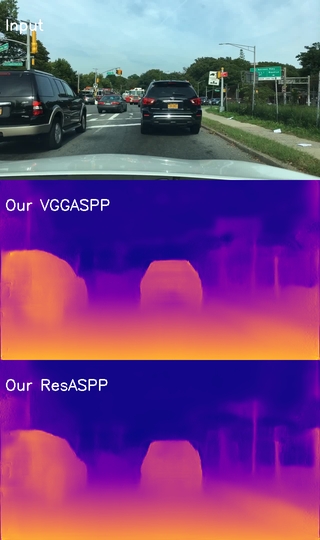}
	\end{subfigure}	
	\begin{subfigure}[t]{.19\textwidth}
		\centering
		\includegraphics[width=1.\textwidth]{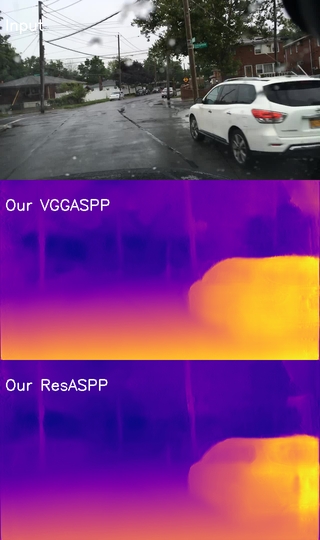}
	\end{subfigure}
	\begin{subfigure}[t]{.19\textwidth}
		\centering
		\includegraphics[width=1.\textwidth]{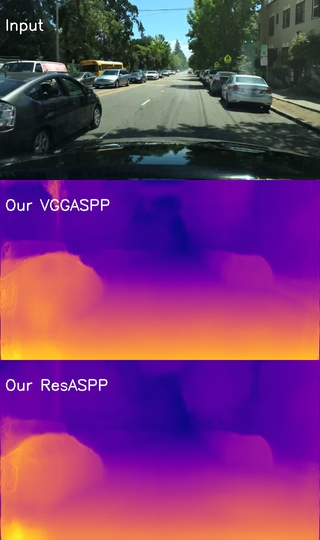} 
	\end{subfigure}
	\begin{subfigure}[t]{.19\textwidth}
		\centering
		\includegraphics[width=1.\textwidth]{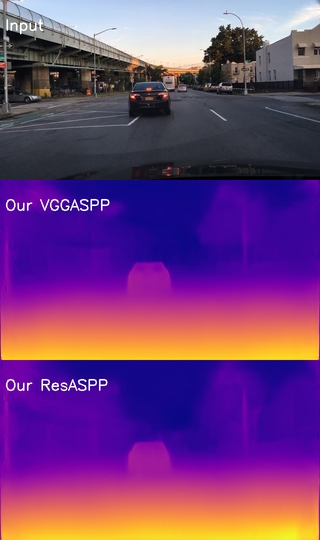}
	\end{subfigure}
	\begin{subfigure}[t]{.19\textwidth}
		\centering
		\includegraphics[width=1.\textwidth]{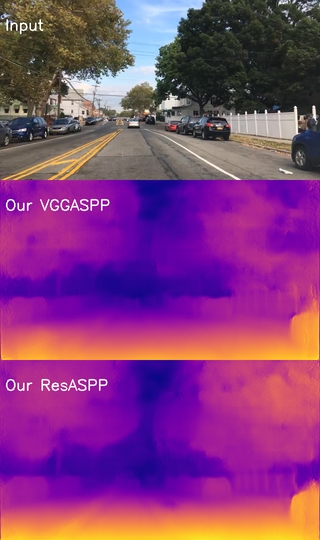}
	\end{subfigure} 
\caption{Deepdrive dataset qualitative results. There are 10 sets of images that we randomly select to show the visualization results.}
\label{fig:lw-eg-sup-demo-deepdrive}
\end{figure*}

We extend our models to experiments on other datasets: Cityscapes \cite{Marius2016cvpr} and Deepdrive \cite{Fisher2018arXiv}. Both these two datasets are outdoor scenes for the auto-driving application. We have used Cityscapes dataset as the pre-train dataset in the training process. We visualize the ten inference results in Fig. \ref{fig:lw-eg-sup-demo-cityscapes}. The Deepdrive dataset is also designed for the auto-driving application with more different types of scenes. There are scenes in different weathers, and it is good for testing the model robustness. The original resolution of Deepdrive dataset is $720\times1280$. We select another ten inference results of Deepdrive dataset shown in Fig. \ref{fig:lw-eg-sup-demo-deepdrive}. The performance of the selected two datasets are consistent as the results on KITTI dataset. The proposed methods can finely capture the object shapes in the test images.

%\bibliographystyle{unsrt}  
%\bibliographystyle{ieee_fullname}
%\bibliography{ms}

\end{appendices}

\end{document}